\documentclass[sigconf,balance=false]{acmart}
\usepackage{popets}

\setcopyright{popets}
\copyrightyear{YYYY}

\acmYear{YYYY}
\acmVolume{YYYY}
\acmNumber{X}
\acmDOI{XXXXXXX.XXXXXXX}
\acmISBN{}
\acmConference{Proceedings on Privacy Enhancing Technologies}
\usepackage{multirow}

\usepackage{enumitem}

\begin{document}

\title{Examining StyleGAN as a Utility-Preserving \\ Face De-identification Method}

%%%%%%%%%%%%%%%% Authors' Info %%%%%%%%%%%%%%%%%

\author{Seyyed Mohammad Sadegh Moosavi Khorzooghi}

\affiliation{%
  \institution{The University of Texas at Arlington}
  \city{}
  \state{}
  \country{}}
\email{seyyedmohammads.moosavikhorzoog@mavs.uta.edu}

\author{Shirin Nilizadeh}
\affiliation{%
  \institution{The University of Texas at Arlington}
  \city{}
  \state{}
  \country{}}
\email{shirin.nilizadeh@uta.edu}

\renewcommand{\shortauthors}{Moosavi et al.}

\begin{abstract}
 { 
Several face de-identification methods have been proposed to preserve users' privacy by obscuring their faces. These methods, however, can degrade the quality of photos, and they usually do not preserve the utility of faces, i.e., their age, gender, pose, and facial expression. 
Recently, advanced generative adversarial network models, such as StyleGAN~\cite{Style-Based}, have been proposed, which generate realistic, high-quality imaginary faces. In this paper, we investigate the use of StyleGAN in generating de-identified faces through style mixing, where the styles or features of the target face and an auxiliary face get mixed to generate a de-identified face that carries the utilities of the target face. 
We examined this de-identification method for preserving utility and privacy by implementing several face detection, verification, and identification attacks and conducting a user study.  
The results from our extensive experiments, human evaluation, and comparison with two state-of-the-art face de-identification methods, i.e., CIAGAN and DeepPrivacy, show that StyleGAN performs on par or better than these methods, preserving users' privacy and images' utility. 
In particular, the results of the machine learning-based experiments show that StyleGAN 0-4 preserves utility better than CIAGAN and DeepPrivacy while preserving privacy at the same level. StyleGAN 0-3 preserves utility at the same level while providing more privacy. 
In this paper, for the first time, we also performed a carefully designed user study to examine both privacy- and utility-preserving properties of StyleGAN 0-3, 0-4, and 0-5, as well as CIAGAN and DeepPrivacy from the human observers' perspectives. 
Our statistical tests showed that 
participants tend to verify and identify StyleGAN 0-5 images easier than DeepPrivacy images. All the methods but StyleGAN 0-5 had significantly lower identification rates than CIAGAN. Regarding utility, as expected, StyleGAN 0-5 performed significantly better in preserving some attributes. Among all methods, on average, participants believe gender has been preserved the most while naturalness has been preserved the least. 
}
\end{abstract}


\maketitle

\section{Introduction}
\label{sec:Introduction}

Posting photos and videos is an integral feature of the design and functioning of the most popular social network sites, such as Instagram and TikTok. 
However, prior research~\cite{li2017effectiveness} has shown users are concerned about uploading photos on  social media, where 53\% of the survey participants have refused to upload a photo on social media in the past because of privacy concerns and 81\% of them would have used obfuscation methods if they had access to them.

Social media platforms do not provide effective functions for obfuscating or de-identifying faces, mainly because they profit from images, as they can be used for identifying relationships and connecting people, and marketing and advertisements. 

In addition, in computer vision, creating datasets or use of high-quality images, which include people, is challenging, as due to data privacy regulations, such as the General Data Protection Regulations (GDPR), people need to consent to the usage of their image data. However, many computer vision tasks, such as person detection, gender, race or emotion detection, or action recognition, do not need to identify the people on the images or videos~\cite{ciagan}. 

Due to these reasons, there is a new trend in the development of face de-identification methods, which try to: (1) \emph{effectively hide the subjects' identity} such that both humans and machines cannot re-identify people; (2) \emph{preserve the realism of visual data}, i.e., makes de-identified subjects look realistic, to be appealing to people, and (3) \emph{preserve the utility of the visual data}, i.e., the people's age, gender, pose, and expression in the data. 
While these three properties are not required in all applications, providing them might encourage users to employ such methods before sharing their photos, and can still enable computer vision researchers and technologies to employ vision tasks that do not need to identify people in the images.  

In particular, existing face de-identification methods have evolved from \emph{image filtering} to more advanced \emph{face de-identification} methods. Image filtering modifies the information using common image filters, such as blurring~\cite{frome2009large,neustaedter2003design, Neustaedter:2006, zhang2006light} or pixelation~\cite{Boyle:2000, Neustaedter:2006, kitahara2004stealth}, which often give unpleasant occlusion. 
The more advanced face de-identification methods either make imperceptible changes to the photo to evade recognition by specific recognition algorithms~\cite{oh2017adversarial, sharif2016accessorize,evtimov2020foggysight,cherepanova2021lowkey,shan2020fawkes}, or substantially modify faces, thus making them unrecognizable for generic recognition algorithms~\cite{sun2018natural,hao2019utility}. 
Some recent work has proposed Generative Adversarial Networks (GANs)~\cite{goodfellow2014generative} for face de-identification, where they generate synthesized objects~\cite{meden2017face, brkic2017know, raval2017protecting, raval2017protecting, mohammadi, wu2019privacy, pittaluga2019learning}. %for the k-same algorithm, Li_2019_CVPR_Workshops, Chen_2018_CVPR_Workshops,sun2018natural, mirza2014conditional
%, which replaces the faces of a given cluster with synthesized faces. 
However, these methods may not preserve the characteristics of the original face. % because they cannot control the image-generation process. 
The resulting faces may have artifacts from inpainting faces of unfitting face poses, expressions, or implausible shapes.

StyleGAN~\cite{Style-Based} is a GAN designed to create high-resolution, realistic but \emph{imaginary} images. Unlike traditional GANs, it can control the features of the generated image or face by style mixing and transferring, in which the generated image inherits the styles or features of an image. 
While StyleGAN has not been originally designed for face de-identification, this paper investigates its effectiveness and robustness as a utility-preserving face de-identification method. 

We show how StyleGAN can be augmented to generate de-identified faces, transferring the original face's features to the de-identified face. We extensively evaluated the privacy of the generated faces through several automatic machine-based re-identification attacks %, i.e., face verification and identification, 
comparing its results with those of two state-of-the-art GAN-based face de- identification methods.  
In addition, we evaluated and compared these methods in terms of their utility-preserving property by employing Face++ on all the de-identified faces. 
To our knowledge, no other work has done such an extensive evaluation of face de-identification methods. Almost all prior studies evaluated the privacy of these methods against either machines~\cite{ciagan, hukkelaas2019deepprivacy} or humans~\cite{li2017effectiveness}. Also, this paper is the first to test their utility-preserving property through detection methods and conducting a user study.

Our findings show that StyleGAN performs better than the state-of-the-art GAN-based de-identification methods in privacy and utility preserving if certain style-mixing levels are used. 
Moreover, since the audience for online photos is human beings, we investigated the human (vs. machine) ability to identify people in the photos and their perception of preserving utility and realism. 
Our findings showed that, in general, human observers are less likely to verify and identify the de-identified images successfully. In addition, StyleGAN models perform on par or even better than CIAGAN. 
Moreover, we found that StyleGAN 0-5 and StyleGAN 0-4 preserve utility attributes, such as naturalness, pose, and expression, more than other models. 
These promising results can inspire the research community to study the properties of StyleGAN and other style-transferring models for developing more advanced utility- and privacy-preserving face de-identification methods.
Thus, this paper has the following contributions: 
(1) Utilized StyleGAN to generate de-identified faces based on the latent vectors of the target's face so that the de-identified faces look different but have the same utility features as the target. 
(2) Implemented extensive experiments to examine the utility- and privacy-preserving properties of StyleGAN for style-mixing levels under different attack models against re-identification attacks, including verification and identification. 
(3) Compared StyleGAN with two recent GAN-based face de-identification methods. 
(4) Carefully designed and conducted a user study to investigate the privacy and utility of de-identified faces from the human perspective. For the first time proposed to use the concept of the police lineup for creating the identification questions. 
(5) To the best of our knowledge, we are the first to compare the utility-preserving property of face de-identification methods in terms of privacy and utility preserving in a human study and using Face++. 
(6) Created high-quality datasets for the community to evaluate face de-identification methods.

\section{Background and Related Work}
\label{sec:relwork}
\subsection{Traditional and K-same Methods} 
Traditional methods, including pixelation, blurring, and masking methods, heavily damage the utility of images~\cite{lander2001evaluating}. %, and faces de-identified by these methods more easily can be recovered than other methods~\cite{DefeatingImage, hill2016effectiveness}. 
Face swapping methods replace a face with another face~\cite{bitouk2008face,perov2020deepfacelab}, trying to preserve privacy and some faces' attributes. 
K-same methods, for example, cluster faces based on some attribute so that similar faces appear in the same cluster, then a de-identified face is generated by, e.g., averaging all the faces in a cluster, which is used for replacing all the faces in the cluster~\cite{newton2005preserving}. 
Some approaches~\cite{bitouk2008face} create a dataset of usually synthetic faces and then propose an algorithm to replace the target face with a face in the dataset with attributes similar to the target's face.  
These de-identified faces suffer in terms of quality and their alignment in the background. Also, since the datasets are small, i.e., not having faces with all attribute combinations, the results might not preserve some of the attributes of the faces~\cite{gross2005integrating, newton2005preserving}.
The state-of-the-art face-swapping methods, including DeepPrivacy~\cite{hukkelaas2019deepprivacy} and CIAGAN's~\cite{ciagan}, use GANs to generate and replace faces with similar attributes trying to fix these problems. They do not need to pre-generate a synthetic dataset, and the de-identified faces can be generated in real-time and look more realistic. 
\subsection{GAN-based Face De-identification} 
Early research on applying GANs for face de-identification began by applying parametric face models~\cite{Replacement}. 
 The GAN-based de-identification methods can be divided into two sets: (1) the methods which rely on conditional inpainting~\cite{hukkelaas2019deepprivacy, sun2018natural, 8578628, ren2018learning}, and (2) the manipulating facial representation methods~\cite{gafni2019live, wang2021infoscrub, li2019anonymousnet, Wang_2021_CVPR, yang2021systematical, li2019identification, cao2021personalized}. 
Nousi et al.~\cite{nousi2020deep} proposed fine-tuning of deep auto-encoders 
to preserve utility and privacy.  
Sun et al.~\cite{sun2018hybrid} used a hybrid model consisting of facial modeling to separate the identity and maintain facial features and inpainting using GANs to add background and make it realistic.  
Agarwal et al.~\cite{agarwal2021privacy} proposed to preserve emotion by extracting facial attributes and feeding the non-biometric vector to the latent vector of an auto-encoder. 
This work used StyleGAN to have a dataset of proxy faces for different facial expressions and poses. The ones with the most similar features replace the target face. Our work, however, examines whether StyleGAN can be used as a face de-identification method.  
Two state-of-the-art de-identification methods, i.e., DeepPrivacy~\cite{hukkelaas2019deepprivacy} and CIAGAN~\cite{ciagan}, leverage conditional generative adversarial networks to remove the identifying characteristics of faces. Since we evaluate and compare the performance of StyleGAN with these methods, these two methods are explained in more detail. 

\subsubsection{DeepPrivacy} DeepPrivacy aims to generate images that preserve the original pose and image background. 
In this method, the background photo (with no face) is fed to the generator in each UNet network layer along with a vector containing pose information. 

\subsubsection{CIAGAN} CIAGAN aims to provide control over the identity of the de-identified face using a vector. Similar to DeepPrivacy, it uses an auto-encoder, which gets background and landmark information as its inputs. The architecture of CIAGAN concatenates a one-hot identity vector to the bottleneck of the generator to ensure generating a face with an identity not present in the dataset. 
Similar to DeepPrivacy and CIAGAN, StylGAN is a conditional GAN that, with our augmentation, enforces the generated face to inherit styles from both the target and an auxiliary face. While DeepPrivacy and CIAGAN attempt to preserve mainly pose, StyleGAN can preserve more attributes, such as gender and expression. 

\subsection{Evaluation of De-identification Methods} 
Unlike privacy-preserving methods that can be defined and measured by mathematical frameworks such as differential privacy~\cite{dwork2006differential} and information theory~\cite{rebollo2009t}, to ensure the privacy of individuals in datasets, face de-identification methods cannot provide privacy guarantees. This is because face de-identification is not per dataset but per image, and the image is high-dimensional data and not aggregated statistics. That is why they are empirically tested against face re-identification attacks. 
\emph{Face re-identification} is a function that attempts to identify the person associated with a face image properly.
Prior research evaluated face de-identification methods by employing a subset of these three approaches: (1) performing privacy vs. utility analysis~\cite{1640608, 10.1007/11767831_15}, (2) studying human or viewer experience~\cite{hasan2018viewer, li2017effectiveness}, and (3) measuring de-identification robustness against adversarial machine learning attacks~\cite{chattopadhyay2021determining,9320277}.
This paper, however, examines StyleGAN through all three approaches.

\section{StyleGAN for Face De-identification}

StyleGAN and its variants are designed as a combination of progressive GAN with neural style transfer, which generates high-resolution imaginary images that look authentic~\cite{karras2020analyzing}. 
\emph{Style transferring} refers to the representation of the content of an image in the style of another using Convolutional Neural Networks ~\cite{8732370}.
Artistic style transfer has been used to create artificial artwork from photographs~\cite{Chen_2021_CVPR}. 
In StyleGAN, style transferring can help transfer some features of a face image, such as pose, shape, and gender, to another imaginary face image. Therefore, we propose to utilize this style transferring property of StyleGAN for generating utility-preserving face de-identification. Here, we provide some background needed to understand the architecture of StyleGAN for face de-identification. 
Here, we provide some background needed to understand the architecture of StyleGAN for face de-identification: 

\paragraph{Latent Vector:} 
In face recognition models, face images are transformed from high-dimensional data (image) to low-dimensional data (latent vector), which uniquely represents the face. Therefore, the face is transformed from the image space to the \emph{latent space}. GANs are the inverse of face recognition models; they transform the latent space into the image space. StyleGAN generates an imaginary but realistic-looking face by mixing the styles of two latent vectors. To use StyleGAN for face de-identification, we propose to mix the latent vectors of the target face and an auxiliary face.

\begin{figure}[t]
\begin{center}
\includegraphics[width=0.8\columnwidth]{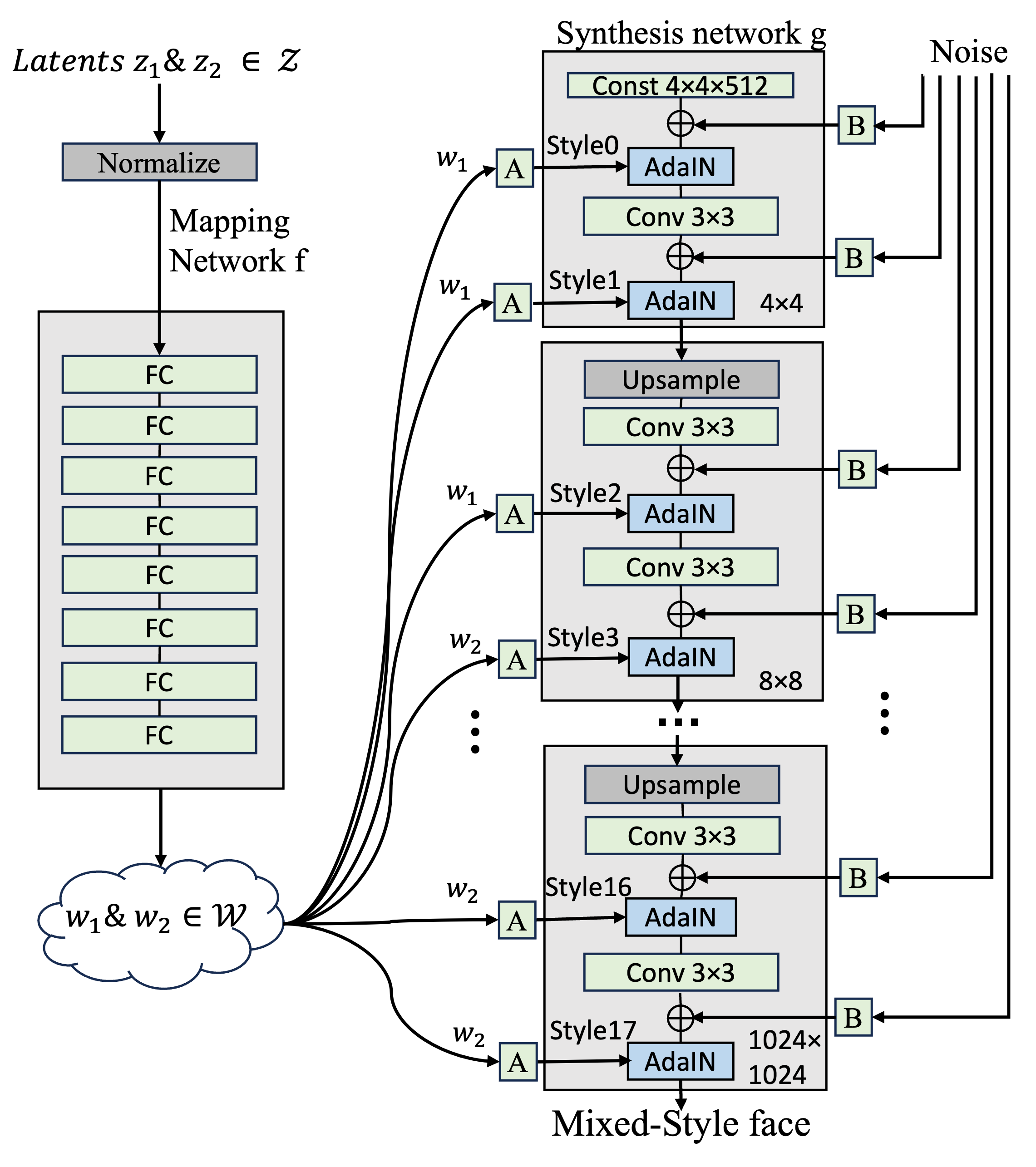}
\caption{Style mixing using StyleGAN~\cite{Style-Based}. In our proposed de-identification method, $z_1$ and $z_2$ are latent vectors for the target and an auxiliary face, respectively. This example shows StyleGAN 0-2 where styles 0-2 are inherited from the target and the rest from the auxiliary face.}
\label{fig:architecture}
\end{center}
\end{figure}

\paragraph{Style Mixing for Face De-identification:} 
As shown in Figure~\ref{fig:architecture}, StyleGAN generates an image based on the latent vector. The latent vector is first mapped into an intermediate latent space, $w$, using Fully Connected Neural Networks (FCs). Then, using the affine transform $A$ and the AdaIN blocks, the style of the latent vector is applied to the feature maps (the output of convolutional layers). The AdaIN block normalizes the feature maps and scales them with the parameters obtained from the affine transform. 
Noise is added using the scale of $B$ to each block. 
StyleGAN-1 and StyleGAN-2 generators consist of convolutional layers with an increased resolution output, starting from a learned constant (4*4*512) and continuing to a high resolution (1024*1024). Therefore, coarser features of the face are composed of earlier layers and finer ones at the last layers.

Transferring the style of the latent input vector to a specific layer transfers the corresponding features of the corresponding face to the generated face. 
Since styles of any latent vector can be transferred to any location, we can consider two latent vectors ($z_1$ and $z_2$) and transfer their styles to the desired locations (e.g., Figure~\ref{fig:architecture} shows styles 0-2 are transferred from $z_1$ and the rest, 3-17, from $z_2$) so that the generated face has certain features, e.g., coarser features of the first face and finer features of the other face. 
Coarser features are more related to utility and finer ones are more related to identity. For de-identification, we propose to generate a de-identified face that inherits the target face's coarser features and the auxiliary face's finer features so that the generated face has the utility features of the target and the identity of the auxiliary face. 

\paragraph{Style Layers:} In StyleGAN-1 and StyleGAN-2, the level of style mixing can be set by $r\in \{0, 1, ..., 17\}$. 
When setting style mixing to $r$, styles $0$ to $r$ are inherited from the target face image, and styles $r+1$ to $17$ are inherited from the auxiliary face image. We refer to style mixing ${0-r}$ when styles $0$ to $r$ are inherited from the target and the rest from the auxiliary face. Similarly, StyleGAN ${0-r}$ refers to the StyleGAN model that de-identifies faces by inheriting styles ${0-r}$ from the target and the rest from an auxiliary face.

\begin{figure}[t]
  \includegraphics[width=0.75\columnwidth]{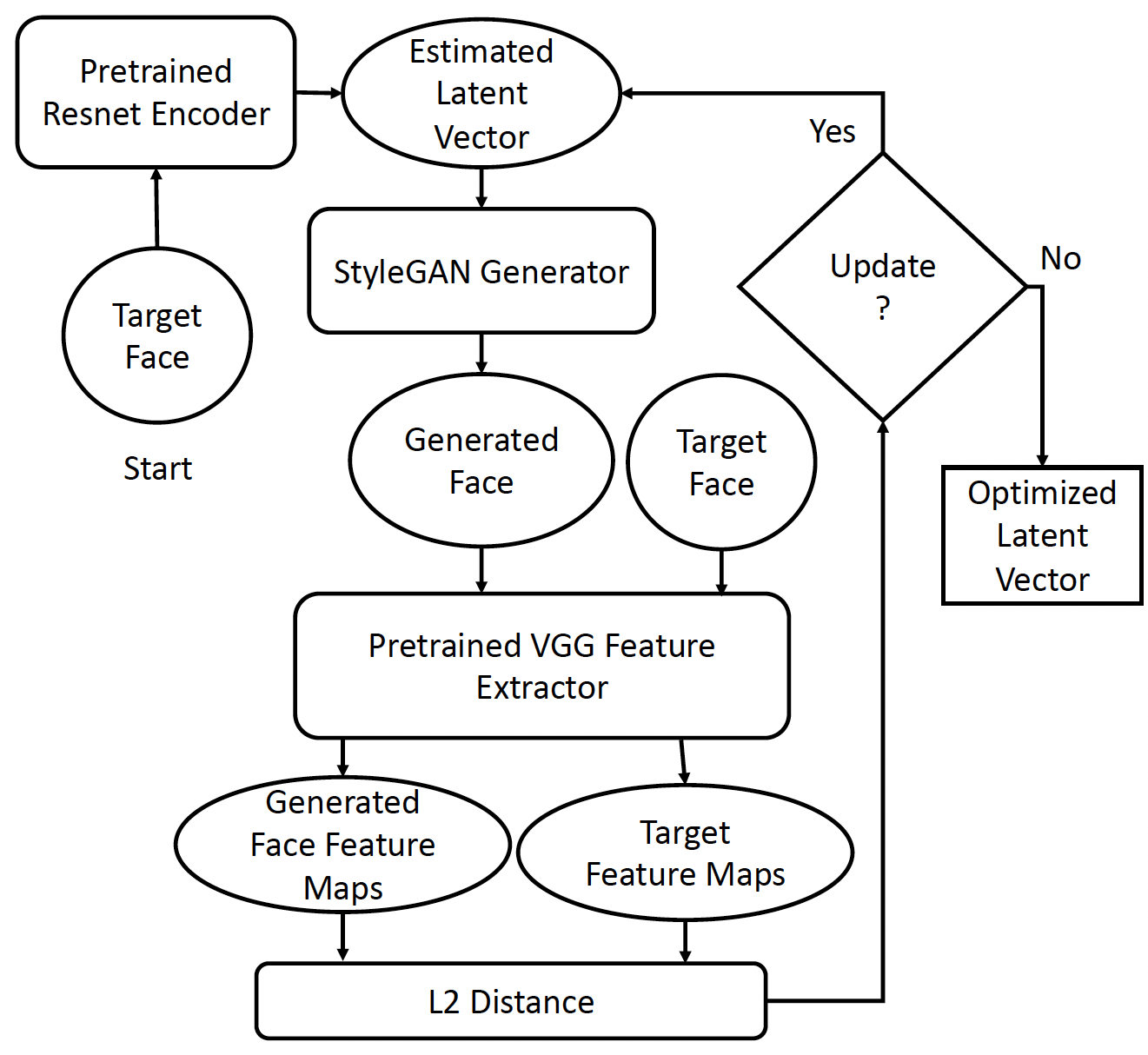}
  \caption{Latent vector generation process. First, a coarse estimation of the latent vector is obtained using a pre-trained encoder. Then, the latent vector is optimized with getting the help of a pre-trained VGG feature extractor.  }
  \label{fig:encoder2}
\end{figure}

\paragraph{Creating Latent Vectors for Face Images:} 
Since the inputs to the StyleGAN are two latent vectors, we need first to obtain the latent vector corresponding to the target face image. Any random latent vector can be used as the auxiliary face. 
As shown in Figure~\ref{fig:encoder2}, the enhanced version of StyleGAN encoder~\cite{karras2020analyzing} is used to generate the latent vectors. This is a ResNet encoder~\cite{image-to-latent} trained on a set of faces and their corresponding latent vectors, which can in return generate an estimated latent vector for a face image. 
To improve this estimate, an optimization algorithm could be used that computes the Euclidean distance between the generated and the target face images as its cost function. However, this approach is too costly. To overcome this limitation, the optimization is enforced on the feature maps instead of the images.  
The feature maps are generated using a pre-trained VGG network and the optimization is performed using L2-optimization. The optimized latent vector is considered a very close estimate of the target latent vector~\cite{image-to-latent}.

%%%%%%%%%%%%%%%%%%%%%%%%%%%%%%%%%%%%%%%%%%%%%%%%%%%%%%%%%%%%%%%%%%%%%%%%%%%%%%%%
\section{Methodology}
\label{sec:methodology}
To examine the effectiveness and robustness of StyleGAN-2 for face de-identification, we first used StyleGAN-2 to create a set of de-identified faces, then we tried identifying these faces by implementing several attacks considering various threat models, ranging from white to gray to black-box adversarial settings. 
We then investigated the quality and utility preserving of de-identified faces generated by StyleGAN-2 by passing them through Face++~\cite{faceplusplus}, a well-known cloud system for face analysis. 
Moreover, we compared the performance of the attacks and Face++ on faces generated by StyleGAN-2 with those generated by two other state-of-the-art face obfuscation methods, CIAGAN~\cite{ciagan} and DeepPrivacy~\cite{hukkelaas2019deepprivacy}. 

\subsection{Threat Model} 
In practice, the use of face de-identification alone might not be enough to protect the privacy of users because attackers, especially human observers, might be able to use contextual features of the images to infer people's identities. 
StyleGAN and other face de-identification methods only modify the face images and not the full images. 
While other face de-identification methods mostly only consider a black-box setting~\cite{chandrasekaran2021face, rosenberg2021fairness, mcpherson2016defeating}, in this paper, we assume that the adversary can have different levels of capabilities and knowledge. We can divide this knowledge into three categories: black box, grey box, and white box. In a black-box setting, the attacker has no knowledge about the method parameters while they know all the parameters in the white-box attack. If they have partial knowledge, the attack is grey-box. 
The adversary's knowledge can be about (1) the StyleGAN style levels that are used for generating the de-identified faces, (2) access to a set of target's face photos, and (3) access to a set of auxiliary faces that are used during the generation of the de-identified faces. 
Table~\ref{table:assumptions} shows different grey-box attack assumptions based on the adversary's knowledge.
In the threat model $m_1$, the attacker is assumed to know about the style level and target photos, however, they do not have access  to the set of auxiliary photos used for de-identification. Therefore, during the attack, a different set of auxiliary pictures are used to create datasets to train the neural network. For another example, in the threat model, $m_4$, the attacker does not know the styles used for de-identification so they would use another mixing style for training.  We then considered a highly capable attacker that has access to a lot of training data. We used the above-mentioned scenarios in our implemented identification attacks. 
If the adversary does not have access to any of this knowledge, then the attack is called \emph{black-box}. 
The table does not include an adversary who has knowledge about all of the three categories because having all that information, the attacker can easily recognize the identity. 
Note that if the target photos are known then the attack is \emph{targeted}, otherwise, it is \emph{untargeted}. 
Threat model m7 is the most plausible because it is less probable that the attacker knows any of the photos and style levels beforehand. In addition, since the auxiliary photo can be generated from any random latent vector, the probability that the attacker learns them is negligible. Moreover, the probability that the attacker has access to the exact image of the target is small. 
%%%%%%%%%%%%%%%%%%%%%%%%%%%%%%%%%%%%%%%%%%%%%%%%%%%%%%%%%%%%%%%%%%%%%%%%%%%%%%
\begin{table}[t]
\centering
\caption{Threat models based on adversaries' knowledge}
\label{table:assumptions}
\resizebox{0.8\columnwidth}{!}{%
 \begin{tabular}{c|ccc} 
 \hline
& \multicolumn{3}{c}{\textbf{Attacker's Knowledge about:}} \\
% \cline{2-4}
Threat Model &  Style levels & Target photos & Auxiliary photos \\  
 \hline 
 $m_1$ & yes & yes & no  \\
 $m_2$ & yes & no & yes  \\
 $m_3$ & yes & no & no\\
 $m_4$ & no & yes & yes\\
 $m_5$ & no & yes & no\\
 $m_6$ & no & no & yes\\ 
 $m_7$ & no & no & no \\
%  \hline
\end{tabular}
}
\end{table}
%%%%%%%%%%%%%%%%%%%%%%%%%%%%%%%%%%%%%%%%%%%%%%%%%%%%%%%%%%%%%%%%%%%%%%%%%%%%%%
\subsection{Data Generation}
\label{data-generation}
\textbf{Dataset of Target Faces:} CelebA dataset was used for our experiment~\cite{liu2015faceattributes}. 
This dataset consists of 10,177 identities with 202,599 face images, i.e., about 20 images for each identity. For each face image, there are 40 binary attributes along with 5 landmark locations. 
This dataset is collected from various sources on the Internet with different levels of quality, enabling us to evaluate the system on photos with various qualities.
Because of process time constraints, we randomly selected a subset of 200 identities with at least 20 images, using an equal number of identities for males and females. Therefore, the total number of images in our dataset is 5,007, approximately 25 images per identity.

\subsubsection{Dataset of Auxiliary Faces}
We used StyleGAN2 to generate auxiliary faces. 
Using the trained StyleGAN2, new faces can be generated based on seed numbers given as input. 
Each seed represents a latent vector. 
Using this capability of StyleGAN2, 400 faces were generated using the first 400 seeds of the pre-trained StyleGAN2. 
Then, 20 of these faces, 2 categories of 10, one for training and the other for validation, were chosen. In each category, we tried to have images from different genders, ages, ethnicity, hair colors, etc. Each of these four categories is used for different experiments. 
For efficiency and to save computation time, we reduced the resolution of images from 1024*1024 to 256*256. 
Still this size is larger than the sizes of all images in the dataset. 

\subsubsection{Generating De-identified Faces} 
The de-identified faces were created by using StyleGAN2 and mixing  5,007 target faces with the 20 auxiliary faces and using 9 style mixing levels, 0-0 to 0-8. 
We generated the de-identified faces using various style mixing levels because they determine the level of information (utility) that is passed from the targets' faces into the de-identified faces. 
For example, in 0-0, the first style, 0, is inherited from the target, and styles 1 to 17 from the auxiliary, so the output image will be almost identical to the auxiliary face, while for 0-8, styles 0 to 8 are inherited from the target and style 9-17 from the auxiliary face. 
In this case, the generated face will have some similar features to the target face. 
There is a trade-off between preserving the utility of a target's face and protecting their identity. 
Our initial tests show that if mixing levels more than 0-6 are used to transfer the target image’s features, the outputs become too similar to the target image, i.e., not preserving privacy. Therefore, in our experiments, to limit the number of unnecessary experiments, we only tested style mixing levels up to 0-8.
Each face was mixed with 20 auxiliary faces generating $20*5,007 = 100,140$ StyleGAN-generated faces for each style. Therefore, there will be $9*100,140 = 901,260$ StyleGAN-generated faces in total. 
In our identification attacks, we use some of these de-identified faces for training and some for validation.

\subsubsection{De-identified Faces by DeepPrivacy and CIAGAN} 
We used two GAN-based utility-preserving face de-identification methods, DeepPrivacy~\cite{hukkelaas2019deepprivacy} and CIAGAN~\cite{ciagan}, to generate de-identified faces for the 5,007 face images. 
While DeepPrivacy could generate de-identified images for all the faces, CIAGAN failed to de-identify 831(~17\%) images due to the face detection failure in its processing phase, where the dataset is prepared for de-identification.  
Figure~\ref{fig:cvpr} shows the generated images of the nine mixing levels (0-0 to 0-8 ) mixed with an auxiliary photo along with de-identified photos obtained from DeepPrivacy and CIAGAN for 4 different target samples chosen from our datasets. The StyleGAN faces change from the most similar to the auxiliary photo (0-0) to the most similar to the target photo (0-8). StyleGAN-generated photos also seem more natural and have better quality than DeepPrivacy and CIAGAN. StyleGAN modifies the background of faces, while other methods do not change it. However, the background can easily be extracted and replaced with that of the generated photo~\cite{image-to-latent}.

\section{Utility and Privacy Analysis}
\subsection{Face Detection} 
Some face de-identification methods cannot even generate a face~\cite{grother2019face}. 
Face quality has also been defined as the usability of an image for recognition~\cite{grother2019face}. 
Therefore, to measure the effectiveness of StyleGAN in generating recognizable faces, we passed all the de-identified images through the Face++ detection module~\cite{faceplusplus}. 
Face++ provides an online free API for face detection, verification, and attribute analysis. The face detection function returns the boundary box of faces in an image, and the face verification function gets two faces as inputs and outputs the probability (confidence) score of these faces belonging to the same identity. The attribute analysis determines different face attributes, including landmarks, age, gender, etc. 
Face++ achieved 99.50 recognition accuracy on LFW dataset~\cite{zhou2015naive} and has been widely used by academics for analysis of images~\cite{sharif2016accessorize, sundararajan2018deep, nilizadeh2016twitter}. 
We found that the Face++ detection module could successfully identify a face in all StyleGAN-generated photos, while it failed for 29 (0.6\%) and 44 (1.1\%) of images generated by Deep Privacy and CIAGAN, respectively. Note that we had 20 (auxiliary photos) * 9 (styles) = 180 times more images for StyleGAN compared to the other two methods, and still all faces were detected.
We manually checked all the photos that Face++ failed to detect a face in them and found that they did not have any face. 
Therefore, we conclude that the quality of StyleGAN-generated faces is higher than CIAGAN and Deep Privacy. 

\subsection{Utility Preserving Evaluation}
Some works have attempted to develop utility-preserving face de-identification methods~\cite{li2019anonymousnet, 8578628, ren2018learning, wu2019privacy}. Transferring non-identity-related features or the utility of the face to the de-identified face increases not only the quality of the photo and its visual appeal but also the de-identified photo can still be analyzed by applications, such as recommendation systems that provide services based on these attributes. 
With advances in image recognition, many  service providers, including Face++~\cite{faceplusplus}, offer online face analysis services to extract face attributes, such as age, gender, emotion, smile, eye status, head pose, mouth status, blurriness, ethnicity, etc. These attributes are also called utility. 
Therefore, we employed some experiments using the Face++ API to check the utility preserving of StyleGAN2. We did not evaluate the effectiveness of StyleGAN in preserving race because this feature does not exist in Face++. 

\subsubsection{Experiments} 
We used 9 datasets of 100,140 faces. The attributes are age, gender, emotion, smile, eye status, head pose, mouth status, blurriness, and ethnicity. Gender could be male or female. Smile is presented with a real number between 0 and 100. Emotion is presented as percentages for 6 different emotion states: anger, disgust, fear, happiness, neutral, sadness, and surprise. 
Mouth status has 4 states: surgical mask or respirator, other occlusions, close, and open, and eye status has 6 states for each eye: occlusion, no glass with an open eye, no glass with a closed eye, normal glass with an open eye, normal glass with a closed eye, dark glasses, and occlusion. The head pose consists of three angles between -180 and 180 degrees: pitch angle, roll angle, and yaw angle. 

\subsubsection{Metrics} 
We computed and compared 
the mean and standard deviation of differences  
for age, blurriness, and smile. For the gender, the ratio of times when the genders are the same is presented. For the emotion and the eye and mouth statuses, the state with maximum confidence value is selected, and the ratio of times the original and de-identified faces have the same chosen state is computed. For the head pose, the mean of differences for each angle is calculated, and then, the mean of the three numbers is presented as the final value for the head pose difference. 

\subsubsection{Results} Table~\ref{table:other_attributes_FPP} shows the results of attribute preserving for different style-mixing levels, CIAGAN, and Deep Privacy. The values of the first 4 attributes, i.e., gender, emotion, eye status, and mouth status, match rates in percent (0-100). 
As expected, our results show that the higher the style mixing is, the more utility is preserved. For example, while gender and emotion are preserved for 47\% and 51\% of the faces when the style mixing is 0-0, these values are 98\% and 84\%  for 0-8. 
This is because when more styles are inherited from the target, thereby having more similar attributes to the target. 
Interestingly, our results show that the attribute values for style 0-3 on-wards are as good as or better than those of CIAGAN and Deep Privacy. For example, gender and emotion are preserved for 81\% and 52\% of de-identified faces generated by CIAGAN, 89\% and 60\% for DeepPrivacy, while 88\% and 73\% for StyleGAN 0-3.  
The values for the other 4 attributes, age, smiling, head pose, and blurring, decrease as style mixing level $n$ increases, i.e., the lower the metric values are, the more the attributes of the original and de-identified faces match. The maximum difference when two attributes are completely different is 100.  
Our results show very small values as small as 5, 6, 2.1, and 9 for age, smiling, head pose, and blurring for faces generated by StyleGAN 0-8. 
Interestingly, the blurring for StyleGAN is much less than that for CIAGAN and Deep privacy, no matter what style mixing is used. Also, we observe that StylGAN with style mixing levels 0-3 to 0-8 has better performance preserving the  smiling and head pose attributes.   
The only attribute not as well-preserved as others when using StyleGAN is age, where CIAGAN and Deep Privacy show better results, i.e., 10, compared to 16 and 13 for StyleGAN 0-3 and 0-4, respectively. 

\begin{table*}[t]
\centering
\caption{Utility preserving of StyleGAN, CIAGAN, and Deep Privacy for 8 attributes using Face++. All StyleGAN models from 0-3 to 0-8 preserve face attributes better than DeepPrivacy and CIAGAN. } 
\label{table:other_attributes_FPP}
\resizebox{0.6\textwidth}{!}{%
 \begin{tabular}{c|c c c c c c c c c c c c} 
  \hline
 \textbf{Method}   & \textbf{gender} &   \textbf{emotion}  & \textbf{eye} &   \textbf{mouth} &  \multicolumn{2}{c}{\textbf{age}} &  \multicolumn{2}{c}{\textbf{smiling}}  & \multicolumn{2}{c}{\textbf{head pose}} &  \multicolumn{2}{c}{\textbf{blurring}}\\
   % \hline
    &  & & &  & M & Std & M & Std & M & Std & M & Std \\
     \hline
   StyleGAN 0-0 & 47\% & 51\% & 76\% & 46\% & 19  & 24 & 42  & 58  & 5.9  & 5 & 10 & 22 \\

  StyleGAN 0-1 & 48\% & 51\% & 80\% & 47\% & 19  & 24 & 43  & 59  & 5.2 & 4.4  & 10  & 22 \\
    % \hline
  StyleGAN 0-2 & 60\% & 62\% & 84\% & 50\% & 17 & 21  & 30  & 47  & 4.3  & 3.8  & 10 & 22 \\
    % \hline
  StyleGAN 0-3 & 88\% & 73\% & 87\% & 65\% & 16  & 20  & 16  & 32  & 3.3  & 3.1  & 10  & 22 \\
    % \hline
  StyleGAN 0-4 & 94\% & 77\% & 89\% & 75\% & 13  & 17 & 12  & 26  & 3.1  & 2.9  & 10  & 22 \\
    % \hline
  StyleGAN 0-5 & 96\% & 80\% & 91\% & 81\% & 10  & 13  & 10 & 23  & 2.8  & 2.7  & 10  & 22 \\
    % \hline
  StyleGAN 0-6 & 96\% & 82\% & 92\% & 83\% & 7  & 10  & 8  & 20  & 2.6  & 2.5  & 9  & 22 \\
    % \hline
  StyleGAN 0-7 & 97\% & 83\% & 93\% & 85\% & 6  & 8 & 7  & 17  & 2.3  & 2.3  & 9  & 21 \\
    % \hline
  StyleGAN 0-8 & 98\% & 84\% & 93\% & 86\% & 5 & 7  & 6  & 15  & 2.1  & 2.1  & 9  & 21 \\
    \hline
   CIAGAN & 81\% & 52\% & 73\% & 72\% & 10  & 12  & 25  & 39  & 3.8  & 4.1 & 26  & 39 \\
    % \hline
   Deep Privacy & 89\% & 60\% & 87\% & 57\% & 10  & 13  & 36  & 46  & 3.6  & 3.7  & 16  & 29 \\
    \hline

\end{tabular}
}
\end{table*}

\subsection{Verification Attack} 
Face de-identification methods are mostly evaluated against two types of attacks~\cite{hao2020robustness}: 
\emph{verification} and \emph{identification} attacks. 
In a verification attack, 
the attacker has a suspect and tries to verify if the de-identified face belongs to that suspect. 
In practice, in this attack, two face images are given to a face recognition system, and the system should report to what extent they belong to the same person. One of the images contains the unknown or the de-identified face, and the other is a face image of the target or suspect.
We used Face++ to implement our verification attacks. 
We define two attack scenarios: (1) when the attacker has the exact image that the target has used for generating their de-identified face, and (2) when the attacker 
has access to another image of the target. 

\subsubsection{Experiments:}  
All the de-identified images with their corresponding non-de-identified images are given to the Face++ \emph{compare} module, which  returns a confidence score for each pair indicating to what extent (from 0 to 100) the identity of the faces is the same. 
The attack is successful if the system has high confidence in two images belonging to the same person. 
In our experiments, we specify a threshold variable, ranging from 0 to 100, and for every image pair, if their obtained confidence score is more than a given threshold, then we label that verification attack a success for the attacker and a failure for the de-identification method, i.e., the de-identification is not successful in obscuring the face, and vice versa. 

\begin{figure}[t]
% \Centering
\begin{center}
\includegraphics[width=0.7\columnwidth]{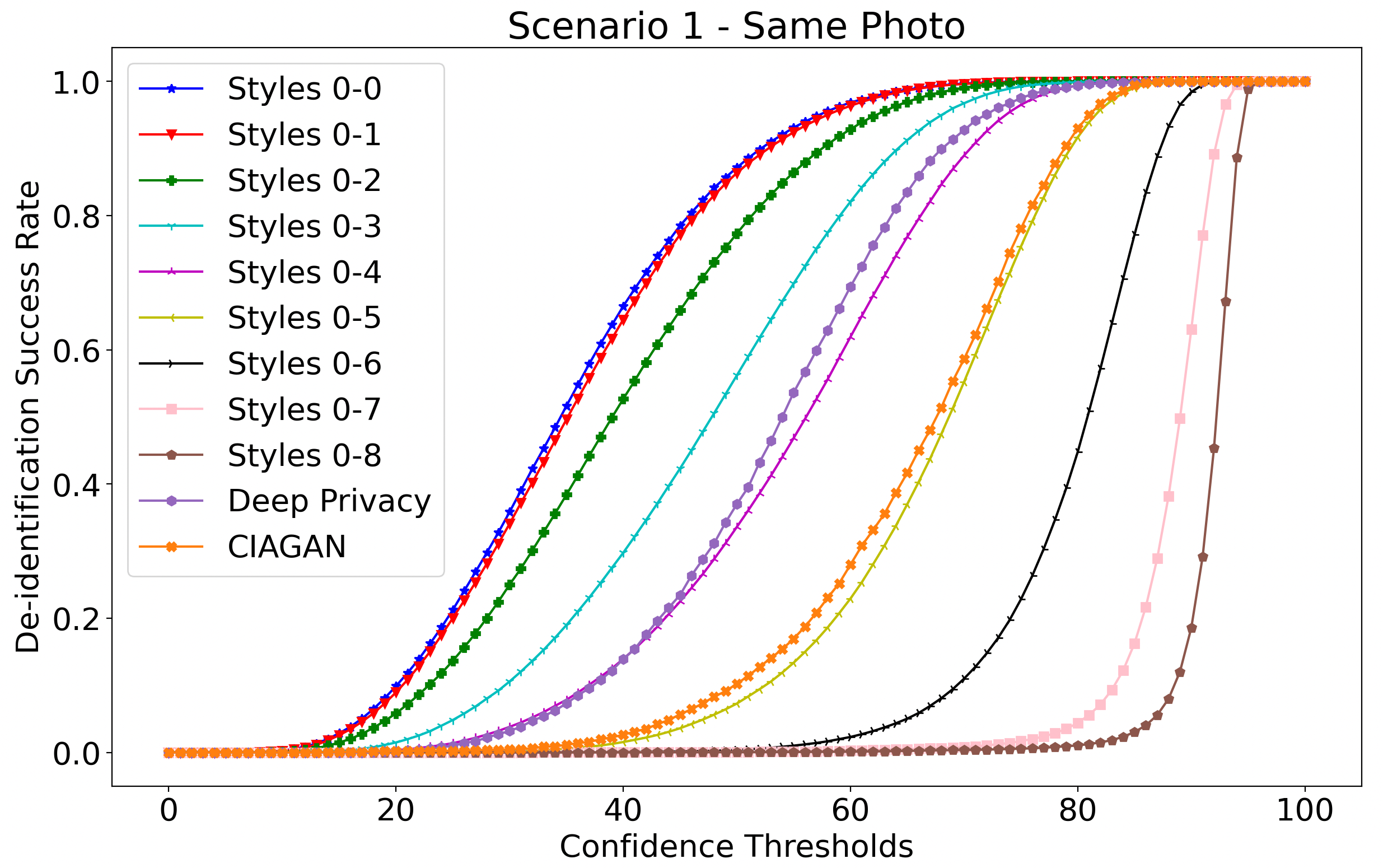}
\caption{Verification Attack results (Scenario 1). The de-identification success rate of StyleGAN 0-3 is far better than DeepPrivacy and CIAGAN. StyleGAN 0-4  performs slightly worse than DeepPrivacy but is still far better than CIAGAN.}

\label{fig:verification1}
\end{center}
\end{figure}

\begin{figure}[t]
% \Centering
\begin{center}
\includegraphics[width=0.7\columnwidth]{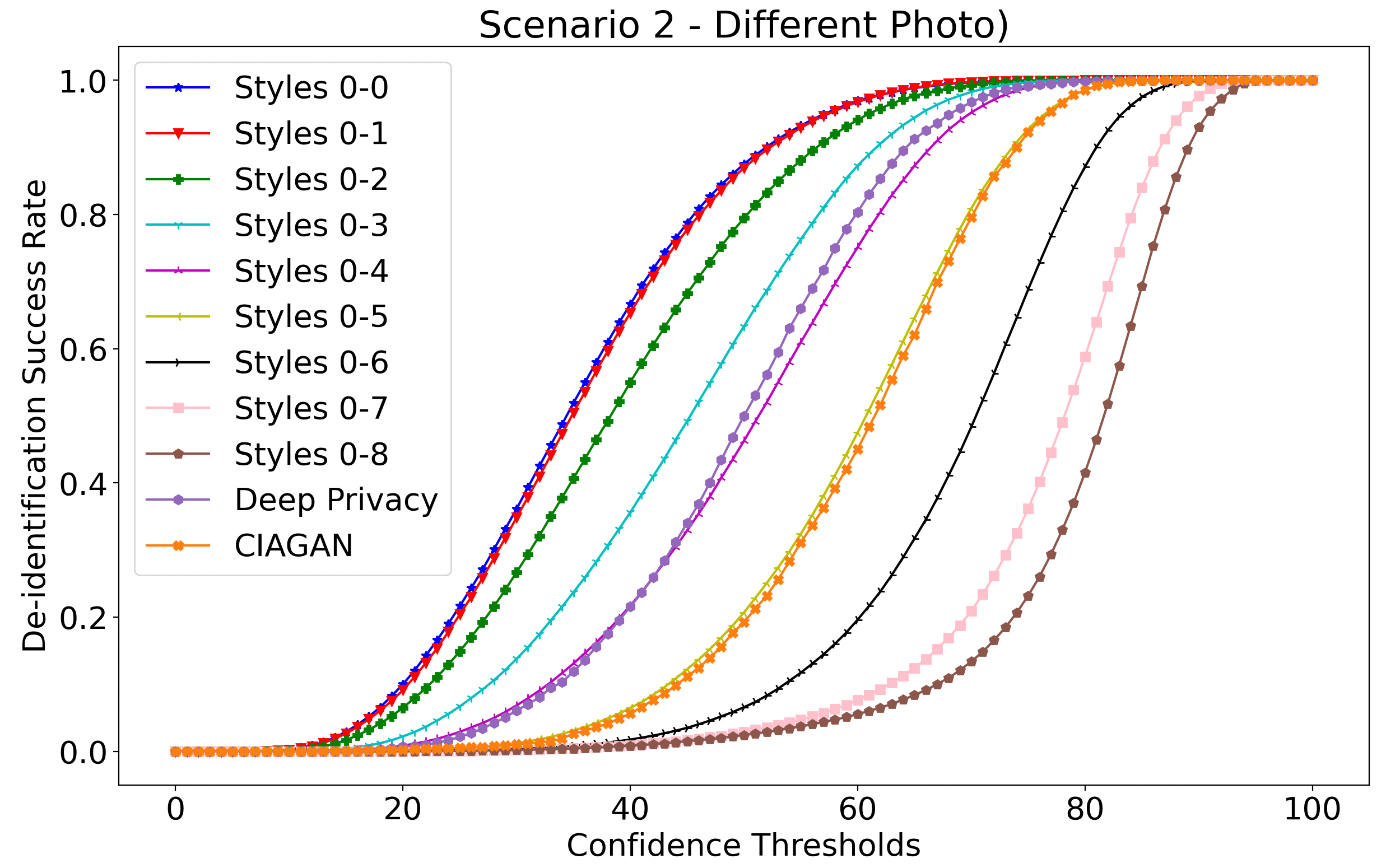}
\caption{Verification Attack results for Scenario 2 when the source photo is different but is the same identity. The results are consistent with  Scenario 1.
}

\label{fig:verification2}
\end{center}
\end{figure}

Figures~\ref{fig:verification1} and~\ref{fig:verification2} show the average de-identification success rate (or 1 - attack success rate) for different threshold values, for StyleGAN with different style-mixing levels, CIAGAN, and Deep Privacy for the first and second scenarios respectively. 
The curves have shifted upwards in scenario 2 compared to scenario 1 confirming our hypothesis that it is harder to re-identify a face when the attacker does not have access to the exact image that the target has de-identified. 
The results for both scenarios are pretty consistent. The higher the threshold, the de-identification success rate is higher. 
De-identified faces generated by StyleGAN 0-0 and 0-1 cannot be re-identified and the de-identified faces generated by higher style-mixing levels are easier to re-identified, especially those generated by 0-7 and 0-8. 
Interestingly, the performance of Deep Privacy is better than CIAGAN and the performance of StyleGAN with style mixing levels 0-3 is far better than both. 
Results for StyleGAN with style mixing levels 0-4 are comparable with those of Deep Privacy and still far better than those of CIAGAN. 
Although de-identification success rates for StyleGAN 0-5 are worse than those of Deep Privacy, they are near those of CIAGAN. 

\subsubsection{Discussion:} 
Remembering the attribute analysis results, StyleGAN with style-mixing levels 0-3 onwards were as good as or better than the state-of-the-art de-identification methods in terms of high utility. 
Here we observe that the de-identification success rates are far better for 0-3. Therefore, based on these results, we argue that StyleGAN, even though was not designed for face obfuscation, performs on par or even better than the state-of-the-art de-identification methods. In addition, because of the ability to choose the mixing levels, the users have the ability to tune the trade-off between privacy and utility. 
For example, if the user is more concerned about privacy, they can choose StyleGAN with style-mixing levels 0-3, and still, the output image preserves adequate utility, or if the user prefers better quality and more utility-preserving de-identified faces, they can choose StyleGAN with style-mixing levels 0-4 and still obtain de-identified images that preserve privacy the same as other state-of-the-art de-identification methods.

\begin{table*}[t]
\centering
\caption{Identification attack results (T: training accuracy, V: validation accuracy) using FaceNet and SVMs. The accuracy values for styles 0-3 and 0-4 in threat model \emph{m7} are lower than those of CIAGAN (31.2\%) and Deep Privacy (30.1\%).}
\label{table:identification}
\resizebox{0.6\textwidth}{!}{%
 \begin{tabular}{c|cc cc cc cc cc cc cc } 
  \hline
 \multirow{2}{*}{\textbf{Method}}   &  \multicolumn{2}{c}{\textbf{m1}} &    \multicolumn{2}{c}{\textbf{m2}} &  \multicolumn{2}{c}{\textbf{m3}} &  \multicolumn{2}{c}{\textbf{m4}} &  \multicolumn{2}{c}{\textbf{m5}} &  \multicolumn{2}{c}{\textbf{m6}} &  \multicolumn{2}{c}{\textbf{m7}} \\
   % \hline
        & T & V & T & V & T & V & T & V & T & V & T & V & T & V \\
     \hline
   StyleGAN 0-0 & 9.8 & .7 & 6.3 & 1.8 & 5.5 & 1.9 & 5 & 12.8 & 9.8 & 16.2 & 4.3 & 5 & 9.2 & 11.1 \\
     % \hline
   StyleGAN 0-1 & 10.2 & .9 & 6.7 & 2.3 & 6 & 2.4 & 5.5 & 13.4 & 10.2 & 17.6 & 5.4 & 6 & 9.6 & 11.5\\
    % \hline
   StyleGAN 0-2 &  15.3 & 2.7 & 98.7 & 2.3 & 10.9 & 6 & 10.5 & 14.2 & 31.7 & 24.7 & 13.1 & 7.4 & 15.3 & 9.9\\
    % \hline
  StyleGAN 0-3 & 31.7 & 11.7 & 98.7 & 3.8 & 27.5 & 18.6 & 26.8 & 25 & 31.7 & 24.7 & 40.8 & 13.7 & 39.1 & 14.8\\
    % \hline
   StyleGAN 0-4 & 50.1 & 27 & 98.7 & 6.7 & 46.4 & 34.2  & 45.7 & 33.7 & 50.1 & 33.3 & 67.6 & 20 & 65.3 & 20\\
    % \hline
   StyleGAN 0-5 & 76.5 & 58.2 & 98.7 & 18.2 & 75.7 & 63.4 & 75.4 & 41.2 & 76.5 & 40.2 & 93.9 & 25.8 & 92 & 26.2\\
    % \hline
   StyleGAN 0-6 & 93.6 & 87.8 & 96.4 & 67.3 & 94.3 & 88.6 & 94.4 & 39.6 & 93.6 & 38.9 & 99.4 & 27.5 & 99.1 & 27.3\\
    % \hline
  StyleGAN 0-7 & 99 & 98.6 & 99.7 & 82.4 & 99.3 & 98.3 & 99.3 & 27.7 & 99 & 26.7 & 100 & 19.5 & 99.9 & 19\\
    % \hline
  StyleGAN 0-8 & 99.6 & 99.5 & 98.7 & 84.4 & 99.8 & 99.3 & 99.8 & 16.2 & 99.6 & 34.2 & 100 & 25.2 & 100 & 24.8\\
    \hline
% \hline
\end{tabular}
}
\end{table*}

\subsection{Identification Attack}
In the identification attack, which is an untargeted attack, the attacker aims to identify a de-identified face by matching it to a person in a set of known people. In other words, identification is a 1-to-N comparison where the goal is to determine if the target is one of the N suspects, whereas verification is  a 1-1 identification.  
We used FaceNet~\cite{schroff2015facenet} to implement the identification attack. FaceNet is a face recognition model, GoogleNet, developed by Google's researchers which reached a state-of-the-art accuracy, of 99.63, by being trained on the Google image set (200 million images for training and 8 million for validation). FaceNet has been used for evaluating face de-identification methods~\cite{rosenberg2021fairness, ciagan}.
FaceNet outputs a distinct feature vector, called face embedding, with a size of 128 for each image. Obtaining the embedding for all the images, then, they can be fed to a classifier, e.g., a Support Vector Machines (SVMs) to perform the identification (i.e., classification). 
We implemented a multi-class SVM classifier with majority voting, i.e., for each possible pair of classes (identities), a binary classifier is trained, and the class of test data is determined by majority voting. 

\subsubsection{Experiments} For this attack, we implemented all the threat models listed in Table~\ref{table:assumptions}.  
The number of images for training and validation changed based on the threat model. For example, if the attacker does not know the auxiliary photos, we considered half of the dataset for training and the rest for validation. 
When the attacker does not know the target photo, we split 70\% of images of the target for training and 30\% for testing. For testing only obfuscated versions of the remaining 30\% were used. 
When the attacker does not know any of the auxiliary photos or the target photo, both conditions were applied. %, i.e., 
For threat models, m4 to m7, the attacker does not know the style-mixing level. In that case, we randomly sampled 20\% of whole images of other styles for validation with other assumptions applied. 
For CIAGAN and Deep Privacy, we considered one threat model in which the attacker has access to 70\%  of the identities and creates de-identified photos for training, and uses the trained model to recognize the identity of the remaining  photos. 
In all the experiments, we used 200 classes (or identities). 

\subsubsection{Results} Table~\ref{table:identification} shows the results of the identification attacks for all style-mixing levels and the seven threat models. 

As expected, the de-identified faces generated by higher style-mixing levels are easier to re-identify, especially those generated by StyleGAN 0-6 upwards. For example, in the threat model m2, the attacker has knowledge about both the style levels and auxiliary photos, the accuracy of the identification attack (validation) for style-mixing levels 0-2, 0-3, 0-4, 0-5, 0-6, 0-7 and 0-8 are 2.3\%, 3.8\%, 6.7\%, 18.2\%, 67.3\%, 82.4\%, and 84.4\%. 
However, there are a few exceptions. For example, in threat models m4 and m5, the accuracy increases until 0-5 and reaches 41.2\% and 40.2\%, respectively, but then it starts decreasing from style levels 0-6, and it gets to 16.2\% and 34.2\% for style levels 0-8. We see a similar but less severe trend for threat models m6 and m7 as well. 

Comparing the validation results for different threat models, we observe that if the attacker has this knowledge about the style levels, i.e., in m1, m2, and m3, the accuracy is much higher compared to other threat models, i.e., m4, m5, m6, and m7. For example, for style levels, 0-6, the validation accuracy for m1, m2, m3, m4, m5, m6, and m7 are 87.6\%, 67.3\%, 88.6\%, 39.6\%, 38.9\%, 27.5\%, and 27.3\%. Note that this assumption that the attacker has knowledge about the style levels is a strong assumption and in practice might not be realistic. 

Running the identification attack on face images of CIAGAN and Deep Privacy, we obtained an accuracy of 31.2\% and 30.1\%, where the threat model is black-box, i.e., equivalent to m7. Comparing their accuracy with those of m7, we see that no matter the style level, it is harder to re-identify faces generated by StyleGAN, as the best verification accuracy for m7, is 24.8\%. Moreover, even no matter the threat model, we observe that the performance of the attack, when StyleGAN 0-3 and 0-4 are used, is better or on par with that of CIAGAN and Deep Privacy. 
\emph{Overall, through our extensive experiments, we showed that StyleGAN is a better de-identification method, especially if style-mixing levels 0-3 and 0-4 are used; 0-4 preserves utility better while 0-3 preserves privacy better.}

\section{Human Evaluation}
We conducted an IRB-approved experiment to evaluate the effectiveness of StyleGAN and other state-of-the-art de-anonymization methods through the eyes of human observers in terms of privacy, utility, and overall quality. 
Particularly, to examine privacy, we test the following hypotheses. Hypotheses \textbf{H1-H3} correspond to the \emph{verification attacks} and hypotheses \textbf{H4-H6} correspond to the \emph{identification attacks}.  
\textbf{H1:} It is less likely for humans to successfully verify a de-identified face. 
\textbf{H2:} It is less likely for humans to successfully verify a de-identified face generated by StyleGAN compared to other GAN-based face de-identification methods. 
\textbf{H3:} It is less likely for humans to successfully verify a de-identified face generated by StyleGAN 0-3 compared to StyleGAN 0-4 and StyleGAN 0-5. 
\textbf{H4:} It is less likely for humans to identify a de-identified face among a set of faces successfully.  
\textbf{H5:} It is less likely for humans to successfully identify a de-identified face among a set of faces when generated by StyleGAN compared to other GAN-based face de-identification methods. 
\textbf{H6:} It is less likely for humans to successfully identify a de-identified face among a set of faces when generated by StyleGAN 0-3 compared to StyleGAN 0-4 and StyleGAN 0-5. 

We examined the utility of the de-identified faces by asking participants whether the original faces and their de-identified versions share any of the following attributes: gender, pose, expression, and age. We also checked for the overall quality of the de-identified faces asking whether they look natural. Our hypotheses for these features are:  
\textbf{H7:} The de-identified faces generated by StyleGAN 0-5 are more likely to preserve the utility features compared to those of other de-identification methods.  
\textbf{H8:} The de-identified faces generated by StyleGAN 0-5 look more natural compared to those of other de-identification methods. 

Moreover, we tried to understand the overall participants' preference about using different face de-identification methods, where 
we de-identified a few samples of images using various face de-identification methods, including \emph{StyleGAN 0-5}, \emph{StyleGAN 0-4}, \emph{StyleGAN 0-3}, \emph{CIAGAN} and \emph{DeepPrivacy}. We analyzed the participants' preference ranking and their justifications.  

\subsection{Experimental Design} 
We used five GAN-based face de-identification methods to examine verification and identification attacks: \emph{StyleGAN 0-5, StyleGAN 0-4, StyleGAN 0-3, CIAGAN,} and \emph{DeepPrivacy}. 
We chose these StyleGAN models because they showed better performance against ML-based attacks. 
We also tested the baseline condition of no de-identification (\emph{as is}). Therefore, in total, we had 6 conditions (5 face de-identification methods plus \emph{as is}).

\subsubsection{Metrics} 
We measured de-identification effectiveness using identification success and confidence, and users' preferences: % via ranking scales. 

\paragraph{Popularity of Face Replacement Compared to Traditional Methods:} We asked ``Assume that you are the woman in the middle of photo X, and you want to upload the photo on social media, but you want to respect the privacy of the others in the photo. You have access to four face obfuscation tools (A, B, C, D) to hide the identity of others. One tool replaces all other faces with new faces, which tries to preserve the age and emotion of people (A), the second one removes the intended people (B), the third one blurs them (C), and the last one replaces their faces with emojis (D). Please order the methods based on your preference ( 1 will be your top choice). What will you choose to use?''

\paragraph{De-identification Effectiveness:}  
We measure the de-identification effectiveness using the same metrics discussed by Li et al.~\cite{li2017effectiveness}: \emph{hit} (the target is present in the choices, and the response is correct), \emph{miss} (the target is present, but the response is incorrect when a wrong person or ``None of above'' is selected), \emph{correct rejection} (the target is absent, and the response is ``None of above''), and \emph{false alarm} (the target is absent, but ``None of above'' is not selected).  

\begin{itemize}[leftmargin=*]
  \item \emph{Verification Success:} We measured verification success by asking, ``Do you identify the below two faces as the same person?'' Two answer choices included ``Yes'' and ``No.'' 
  
  \item \emph{Identification Success:} We measured identification success by asking, ``In the people listed below (a-d), please identify the person indicated by X.'' Five answer choices included four face photos (a-d) and ``None of above.''
  
  \item \emph{Confidence:} After each Verification and identification, we measured confidence using the question, ``How confident do you feel about your answer?” Participants rated their response on a scale from 1 `Completely unconfident' to 7 `Completely confident,' where a higher score meant more confidence. 
\end{itemize}

\paragraph{Utility Preserving Effectiveness:} 
We measured each de-identification method's effectiveness in preserving  the utility attributes by asking ``What attributes of these two faces are the same?'' Five answers were shown as check-boxes and included \emph{Gender}, \emph{Face pose (position)}, \emph{Expression}, \emph{Age (Below ten years difference)}, and \emph{Looking natural (choose if they both look natural)}.  

\paragraph{Looking Natural or Realistic:} 
We measured and compared the quality of generated images by the five de-identification methods, in terms of realistic looking,  by asking `` In the picture below, X is a real face. If you want to hide the identity of X while having a realistic-looking or natural face, what is your order of preference? (1 will be your best and five will be your worst choice).’’ 

\paragraph{Overall Preference:} We measured overall preference by asking ``If you want to hide the identity of face X by using one of the faces below, which one will be your top choice?'' The answers are all de-identified versions of face X using each of the five de-identification methods. 
The answers to the two previous questions might be different because some people might prefer methods that do not generate natural-looking images.

\subsubsection{Selecting Faces} 
We selected faces from the CelebA dataset. CelebA consists of photos of more than 10,000 celebrities. Therefore, knowing these celebrities might make it easier for the participants to identify faces, leading to higher identification success and a lower bound for obfuscation success. 
In \emph{verification questions}, when the target is present, two face images of the target are selected, one gets de-identified, and both pictures are shown to the participants. 
In the second scenario, when the target is absent, one image of the target is de-identified and shown along with another person's photo. 
In \emph{identification questions}, when the target is present, another photo of one of the people in the choices is selected and de-identified or untouched (in the \emph{as is} condition) and shown as X. 

In \emph{verification questions} when the target is absent and in \emph{identification questions}, we add images of other identities to the questions. 
This selection can impact the results, e.g., suppose images are very different from each other and the target. It could be easier for the participant to select ``No'' in verification and ``none of the above'' in identification questions. 
Therefore, we tried using faces with similar attributes, inspired by the \emph{Police Lineups}~\cite{policelineup}.  
In a police lineup process, a crime victim or witness's putative identification of a suspect is confirmed to a level that can count as evidence at trial. 
In this process, the suspect, along with several ``fillers'' or ``foils''—people of similar height, build, and complexion stands side-by-side, facing and in profile.
Like Police Lineup, in identification questions, we showed the photo of person X and asked if participants could identify X's face in a line of four other faces \emph{with similar face attributes}. Similarly, we chose two faces with similar face attributes in the verification target-absent questions. 

\paragraph{Face Clustering Algorithm:}  
Since we randomly selected faces to be added to our questions, we also tried to cluster faces so that we could automatically identify faces of similar attributes to be used along with the target face or in choices. 
To find face images with similar attributes, we applied clustering on all images of the CelebA dataset using the 40 attribute labels provided with the images, including ``Arched\_Eyebrows,'' ``Young,'' ``Bald,'' ``Blurry,'' ``Double\_Chin,'' ``Wearing\_Necklace,'' etc. 
In addition, participants might identify people just based on having the same pose, age, and emotion. Therefore, to minimize the impact of these factors on participants' decisions, we also used Face++ to obtain the attributes for images on each cluster and only maintained those with similar attributes. 
Our criteria regarding age were that the age difference between faces of a cluster should be lower than 10, while for the pose, the absolute difference of different pose angles from the cluster's mean should not be more than half of the standard deviation.
Since we needed four choices for each identification question, we excluded clusters with less than 4 identities.  The final dataset contained 37 clusters, 24 female ones, and 13 male ones. 
This double clustering made sure that faces in a cluster have very high similarity in face attributes. 
In the identification target-present questions, for choices, we used an image of the target and images of three other people from the cluster that includes the target. We used another image of the target for de-identification because de-identifying the same image results in an image with the de-identified face but with the same background as the original image, which could give a hint to the participants. To minimize the impact of this factor on the participants' decisions, we used another image of the target for de-identification and presented it as image X in the questions. 

\subsubsection{Survey Design} 
The survey was implemented on QuestionPro~\cite{quesionpro}. The link to the survey was posted on Amazon Mechanical Turk~\cite{mturk}. 
Our participants were required to have the following qualifications: (1) being residents of the US, (2) being with at least 18 years of age, and (3) having a HIT approval rate of at least \%95 in more than 1000 HITs. 
Having the above requirements, the Amazon workers will be directed to our website, providing instructions on how to do the survey along with the link to the survey on QuestionPro. 
The survey consists of 3 sections: (1) consent form, (2) demographic questions, and (3) main survey questions. 
\paragraph{Consent Form:} 
It contains the contact information about the investigators, the study's goal, IRB approval, research procedure, the collected data, possible benefits and risks, compensation, confidentiality, and the consent statement and question. They are asked if they want to continue the survey. They should select ``yes, I would like to continue to the survey,'' otherwise, their survey is ended. 

\paragraph{Demographic Questions:} After consenting, the participants answered demographic questions about age, gender, education, and race. The majority of the participants are male (66\%), then female (33\%), and others (0.95\%). In order, about 79\%, 8\%, 5\%, 4\%, 4\%, and 0.95\% of participants are white, Asian, Black or African American, Hispanic or Latino, American Indian or Alaskan Native, and from multiple races.  The age distribution is 40\%, 25-34, 37\%, 35-44, 10\% 45-54, 10\% 55-64, 2\% 18-24, and 2\% above 64. 
Most participants (about 61\%) have bachelor's degrees, and 14\% and 9.5\% have master's and high-school degrees. 
Others have associate degrees (6.5\%) and college without degrees (8.5\%).  

%%%%%%%%%%%%%%%%%%%%%%%%%%%%%%%%%%%%%%%%%%%%%%%%%%%%%%%%%%%%%%%%%%%%%%%%
\paragraph{Main Survey Questions:} The main part of the survey includes four clusters of questions: (1)~verification, (2)~identification, (3)~utility-preserving, and (4) method preference. 
For each condition, we created ten questions, with five females and five males as target faces. For example, to test StyleGAN 0-3 against the verification attack, when the target was present, ten questions using different targets were created, and only one of them was shown to each participant randomly. 
We included three attention-check questions randomly throughout the survey. We carefully designed these attention checks to be fair. For example, since in our survey, participants observe many verification questions with the exact question text, they might not read the questions but answer them by just investigating the images. 
Therefore, in attention-checks, instead of asking them to choose a specific choice, we put images with obvious answers and expected to obtain correct answers.  
For example, for a verification question, we showed the same image for both the original and de-identified images, expecting an honest user to verify that the images belong to the same person.  
Having answered all the questions, they were given a six-digit number to write down and enter in the box provided on Amazon Mechanical Turk to be compensated. 

\subsection{Results}
The experiment was completed by 161 participants. We excluded the data of 8 participants who failed at least one attention check question. 
Therefore, the final sample size is 153. 

\subsubsection{Compared with Traditional Methods} 
\begin{figure}[t]
  \includegraphics[width=0.65\columnwidth]{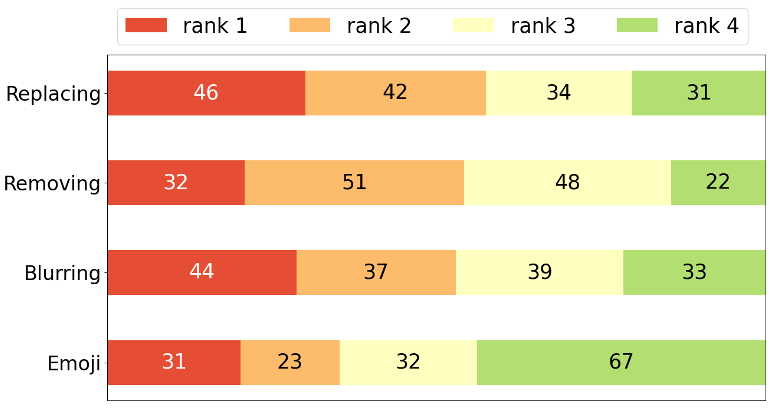}
  \caption{Ranking of obfuscation methods. Participants favored face replacement the most and using emojis the least.}
  \label{fig:tool_ranks}
\end{figure}

We first examined if participants would find utility-preserving face de-identification methods compelling and would consider using them in the presence of other more common methods, such as blurring, using emojis, or removing the faces (see Figure~\ref{fig:survey_tool_rank.png}). 
We asked the participants to put themselves in a situation where they want to upload a photo of themselves at a convention on social media, but they want to respect the privacy of others in the photo. They have access to four face de-identification tools to hide the identity of others, including (A) a face replacement method that preserves the age and emotion of people, (B) a method that removes faces, (C) a blurring method, and (D) replacing faces with emojis. They could order the methods based on their preference.  
This question was followed with another explanation question asking to explain the reasons for their top favorite method. The results of the ranking are shown in Figure~\ref{fig:tool_ranks}. It seems that face replacement and blurring were the most favorite, with being chosen 46 and 44 times as the top favorite methods, respectively, while body removal and emoji replacement were the least favorite ones, with being chosen 31 and 31 times as the best method. We also calculated the means and standard errors of the means for each of these four methods. 
Face replacement had the lowest mean (higher ranks) of 2.327 (std err= 0.09), and blurring was in second place with a mean of 2.399 (std err= 0.09). Body removal and emoji replacement were the third (mean= 2.392, std err= 0.079 ) and last (mean= 2.882, std err= 0.095). 
Even though on average, face replacement receives a better ranking than other de-identification methods, many participants chose other methods over face replacement. This might show that people have different criteria for selecting these methods. Next, we analyzed the qualitative reasoning provided by the participants.   

\paragraph{Reasons for the Top Favorite Method:} 
We investigated the reasons given by participants about why they chose a particular method as their top method. 
We applied the open coding process~\cite{glaser2017discovery} to categorize reasons. 
Following this process, we define new categories until no new categories emerged. To improve the quality of the categories, we used an iterative process~\cite{corbin1990grounded} so that new categories were added or existing ones were reorganized. 

\paragraph{Face Replacement:} About one-third of the explanations provided by the 46 participants, who selected face replacement, were not related or clear. Ten participants stated that they chose this method because of its quality and clarity. Being normal, natural, or not attention-grabbing was mentioned 7 times. Similarity to the original photo, fewer editions
%, and distortion 
were also noted 7 times. 
%The remaining less frequent reasons include being legal, wise, easier to use, and more privacy. 
On the other hand, the reasons of those who disliked the method included ``The idea is creepy,'' ``disrespecting other people in the photo,'' ``It’s a lie,'' ``less aseptically pleasing,'' ``The faces look too similar to the original,'' ``Making an alternative-reality  is creepy.''

\paragraph{Face Removal:} Twenty-five participants, among the 32 participants who selected face removal, explained their reasons properly. Nine stated that this method can preserve privacy very well as it completely removes the person, as in ``It’s like that no one has ever been there.'' 
Seven participants noted that the picture was aesthetically more pleasing by stating ``It looks cool,'' ``nice,'' or ``pleasant.'' 
Four noted that it was natural and not noticeable. Three indicated that it was clear and clean looking, and another three participants said the picture is closer to the original picture and retains the form. 
On the other hand, some participants believed that removal “causes confusion” or “shifts the attention to what is missing” because of the background or “doesn't capture the scene.” 

\paragraph{Blurring:} Of 43 participants who chose blurring, 30 provided proper reasoning. A variety of reasons was given for this method. The most frequent answer was related to the aesthetics of the images, which were mentioned eight times in phrases like “pleasant,” “attractive,” “interesting,” “cute," and “best looking.” 
The second primary reason was “less distracting or noticeable,” “not affecting or altering the image,” “not drastic,” which were mentioned six times. One important reason was that it transfers the message to viewers of what we have done or want to hide from them. Other reasons included “professional” ( 3 times), “common and normal” (3 times), “not weird,” “ethical,” “respectable,” “least awkward,” etc.

\paragraph{Emojis:}  The emoji method got 31 votes with 22 proper explanations. Privacy was mentioned eight times. Three participants stated they just liked it the most. 
They said that emojis were big and obvious and made it clear that we did that for privacy reasons. Other reasons included being “funny” or “cute,” “simple,” “no tool,” “no editing,” showing “emotion” or “expression,” “natural,” “not odd,” “not removing anybody from the image,” “looks like a person,” and “suitable for blanc faces.” 
Emoji was also the most disliked method: 5 participants stated that it is “childish” and “unprofessional” and “disrespecting the photo and people in it.” Other terms that were mentioned regarding using emojis are “too distracting.” “Silly,” “goofy looking,” “ridiculous looking,” and “gets all focus.” 

In summary, participants gave varied reasons to support their choices. One reason, considered negative for a participant, might be positive for another. For example, one stated that emojis get all the focus and attention of the image. At the same time, the other said that emojis convey that identity is hidden due to privacy. 
We believe that some participants might have been confused about face replacement because they might assume the replaced faces belong to real identities while they could be imaginary ones.  
Moreover, the participants might not be aware that traditional methods such as blurring are not effective against machine learning-based attacks. 

\subsubsection{Obfuscation Effectiveness:} 
\begin{figure}[t]
 \includegraphics[width=0.65\columnwidth]{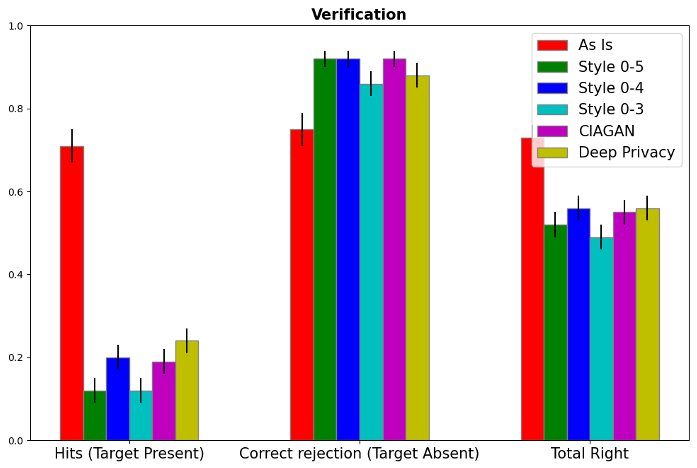}
\caption{Verification success evaluation. All the methods perform better than ``As Is'' especially in the hit rate. 
}

\label{fig:verification}
\end{figure}

We classify \emph{verification success} using three categories: among all cases, among questions where the target is present, and among questions where the target is absent. Having a lower identification success rate means that the de-identification method is more effective and the attacker is unable to identify faces. 
\paragraph{Verification Success:} 
Figure~\ref{fig:verification} shows verification success, correct rejection, and total correct for As Is and the 5 de-identification methods.
As you can see, there is a significant difference between the hit rate and total correct of As Is (hit= 71\%, total 73\%) and those of all the de-identification methods, including StyleGAN 0-5 (hit= 12\%,  total 52\%), StyleGAN 0-4 (hit= 20\%,  total 56\%), StyleGAN 0-3 (hit= 12\%,  total 49\%), CIAGAN (hit= 19\%,  total 55\%), and DeepPrivacy (hit= 24\%,  total 56\%). 
However, we observe that in contrast, the correct rejection is lower for As Is compared to all other de-identification methods. 
The reason can be that we intentionally chose images with very similar attributes and it has made it less trivial for the participants to determine a target is absent. 
We also ran statistical tests to examine \emph{H1}. Since the questions are provided to the same participants, we used logistic mixed-effects model while clustering on participant ids. 
Running two separate logistic mixed-effects models on hits and all cases shows that the success rate of \emph{As Is}, in these two conditions, is higher than all de-identification methods (all $p<0.0001$). 
Running the logistic mixed-effects model on correct rejection shows that the success rate of \emph{As Is}, is lower than all de-identification methods (all $p<0.0001$). 
\emph{Therefore, these results support our first hypothesis that it is less likely for humans to successfully verify the identity of a de-identified image.} 
To examine H2, we tested six t-test hypotheses with Bonferroni Correction, comparing the verification success rate of three StyleGAN models with DeepPrivacy and CIAGAN. We chose the significant level of 0.05 for the t-test. Bonferroni Correction changes the significant level to 0.0083. 
The results show that there is a statistically significant difference between the success rates of StyleGAN 0-5 (hit rate = 0.24) and DeepPrivacy (hit rate =0.12) for the Hit Rates ($p<0.007$). However, there was no statistically significant difference between the success rates of StyleGAN 0-3 and CIAGAN/ DeepPrivacy, StyleGAN 0-4 and CIAGAN/ DeepPrivacy, or StyleGAN 0-5 and CIAGAN. This shows that these models have a similar performance and H2 is only partially supported. 
To examine H3, we tested two t-test hypotheses with Bonferroni Correction, comparing the verification success rate of three StyleGAN models.  Bonferroni Correction changes the significant level to 0.025. 
The results show that there was no statistically significant difference between the success rates of StyleGAN 0-3 and that of 0-4 and 0-5. 
Therefore, H3 is not supported. 
% %%%%%%%%%%%%%%%%%%%%%%%%%%%%%%%%%%%%%%%%%%%%%%
\begin{figure}[t]
  \includegraphics[width=0.65\columnwidth]{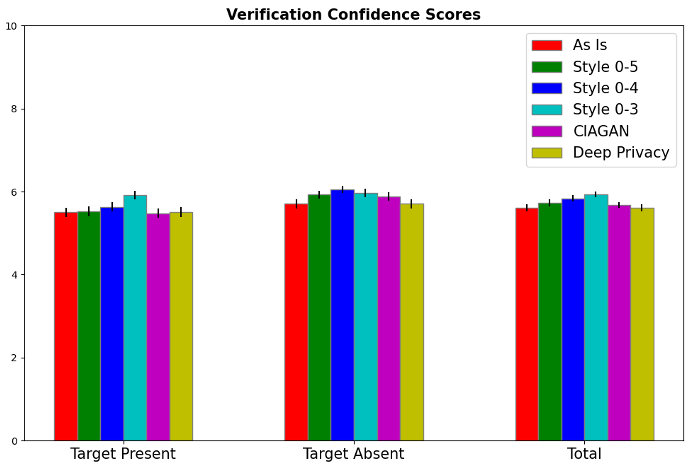}
  \caption{Average confidence scores. We did not find significant differences in the confidence level of survey participants.}
  \label{fig:verification_confidence}
\end{figure}

\paragraph{Confidence Scores:} Figure~\ref{fig:verification_confidence} shows the means of confidence scores for Target-Present and Target-Absent and their combinations. On average, for all the questions and methods, the participants stated that they were confident when verifying images since the means are close to 6, with 7 as "Highly Confident." 
By performing statistical t-tests using Bonferroni correction (significant threshold = 0.0033) on all cases, we found significant statistical evidence that the participants are more confident when verifying CIAGAN images compared to As Is images ($m1=5.94$, $m2= 5.61$, $p<0.001244$), similarly when answering StyleGAN 0-3 questions compared to DeepPrivacy ones ($m1=5.94$, $m2=5.61$, $p<0.00185$). 
Running the t-tests for the target-present or target-absent questions, we found a statistically significant difference between the confidence score of StyleGAN 0-3 and CIAGAN when the target was present, where people were more confident for StyleGAN 0-3 ($m1=5.92$, $m2=5.48$, $p<0.0033$). 
We examined if participants were more confident when correctly answering the questions. Even though the difference between the mean values is small (about 0.4), we surprisingly observed that participants were more confident when their responses were inaccurate ($m1=5.81$, $m2= 5.40$, $p<0.0001$).

\begin{figure}[t]
  \includegraphics[width=0.65\columnwidth]{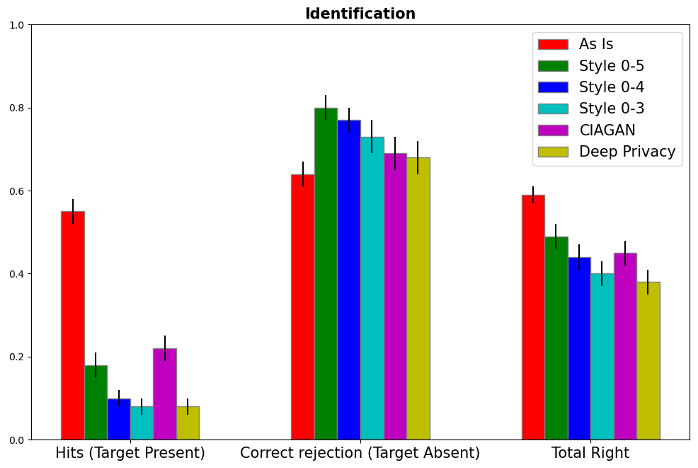}
  \caption{Identification Success. Similar to verification, when the target is present, the hit rate of ``As Is'' is higher than in de-identification methods.}
  \label{fig:identification}
\end{figure}

\begin{figure}[t]
  \includegraphics[width=0.65\columnwidth]{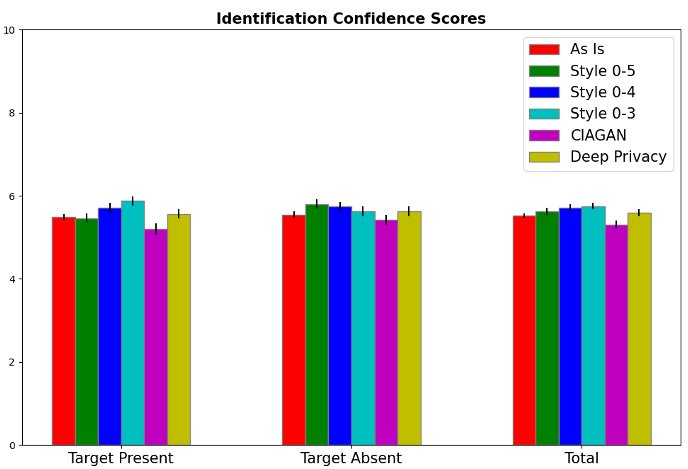}
  \caption{Average confidence scores. There are no  significant differences in the confidence levels.}
  \label{fig:identification_confidence}
\end{figure}

\paragraph{Identification Success:} 
Figure~\ref{fig:identification} shows the identification success rates for different obfuscation conditions for the identification questions. Similar to the verification, when the target is present, the hit rate of \emph{As Is} is far higher than other de-identification methods. The chance hit rate here is $\frac{1}{5}=0.2$, since participants have 5 options to choose from. The hit rate of \emph{As Is} is far above this chance hit rate while that of the de-identification methods is near or blow it. The correct rejection rate again is lower for \emph{As Is}. 
To examine H4, we employed logistic mixed-effects model while clustering on participant IDs. 
Running three separate logistic mixed-effects models on hits, correct rejection, and all cases. 
The results of the model on hits and all cases show that the success rate of \emph{As Is} is higher than all de-identification methods (all $p<0.001$). 
\emph{Therefore, these results support H4 such that it is less likely for humans to successfully identify a de-identified face among a set of faces.} 

To examine H5 and H6, we tested eight t-test hypotheses with Bonferroni Correction, comparing the verification success rate of three StyleGAN models with DeepPrivacy and CIAGAN. 
% We chose the significant level of 0.05 for the t-test. 
Bonferroni Correction changes the significant level to 0.0062. 
The results show a statistically significant difference between the success rates of StyleGAN 0-4 and 0-3 and CIAGAN for hits ($p<0.005$), where it is easier for humans to identify faces generated by CIAGAN. 
We found a statistically significant difference between the success rates of StyleGAN 0-5 and DeepPrivacy for all cases and hits ($p<0.0004$), where it is harder for humans to identify faces generated by DeepPrivacy. We found no statistically significant in other models.  
Therefore, H5 is partially supported, and it shows that all StyleGAN models perform as good or better than CIAGAN and DeepPrivacy, except faces generated by DeepPrivacy compared to faces generated by StyleGAN 0-5 are harder to be identified. 
Interestingly, the results show that there is no statistically significant difference between hit rates of StyleGAN 0-3 and StyleGAN 0-4, and StyleGAN 0-3 and StyleGAN 0-5. Therefore, H6 is not supported. 

\paragraph{Confidence Scores:} 
Figure~\ref{fig:identification_confidence} shows mean confidence scores and their standard errors for the identification questions. The average rates for all the scenarios are high (near 6 on a scale of 1-7) indicating that the participants felt confident when answering  the identification questions. Similarly, performing a statistical t-test using Bonferroni correction for different hypothesis combinations, we found that the total confidence scores of StyleGAN 0-4 and 0-3 were significantly higher than that of CIAGAN ($p<0.0002$). There was only one other significant difference where participants felt more confident when they answered StyleGAN 0-3 questions compared with CIAGAN questions when the target was present ($p<0.00001$). 
Similar to verification, participants felt more confident when their answers were inaccurate ($m1 = 5.79$, $m2=5.02$, $p< 2.2e-16$).

\begin{figure}[t]
  \includegraphics[width=0.65\columnwidth]{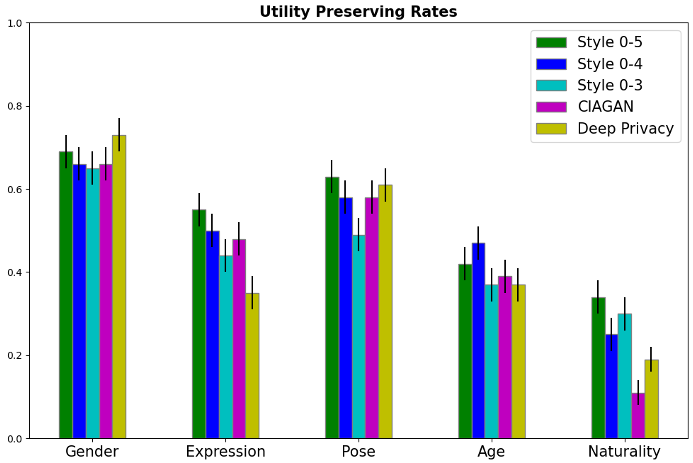}
  \caption{The utility preserving rates based on human judgment. On average, gender and pose are preserved the most and naturality the least. }
  \label{fig:utility}
\end{figure}

\subsubsection{Utility Preserving} 
We created one question for each method asking if the original photo and its obfuscated version share the same gender, expression, pose, age (less than 10-year difference), and naturalness. 
We randomly picked 20 paired sample pictures, 10 for males and 10 for females. One random pair was shown to a participant. Participants could select none or all of the facial features. 
The attribute rates for different methods are shown in Figure~\ref{fig:utility}. The attribute rate here is defined as the number of participants who checked a particular attribute for a certain method divided by the total number of participants. Among all methods, on average, the participants believe gender has been preserved the most while naturalness has been preserved the least. 
To examine H7, we employed logistic mixed-effects model for each of the utilities, comparing the performance of StyleGAN 0-5 and other face de-identification methods.  
Results show that the faces generated by StyleGAN 0-5 are more likely to preserve  
 \emph{expression} compared to DeepPrivacy ($p< 8.41e-05$), \emph{pose} compared to StyleGAN 0-3 ($p<0.002$), and \emph{naturalness} compared to CIAGAN ($p<2e-07$) and DeepPrivacy ($p<0.0006$).  
Therefore, H7 is supported. 

\subsubsection{Ranking De-identification Methods} 
\begin{figure}[t]
  \includegraphics[width=0.65\columnwidth]{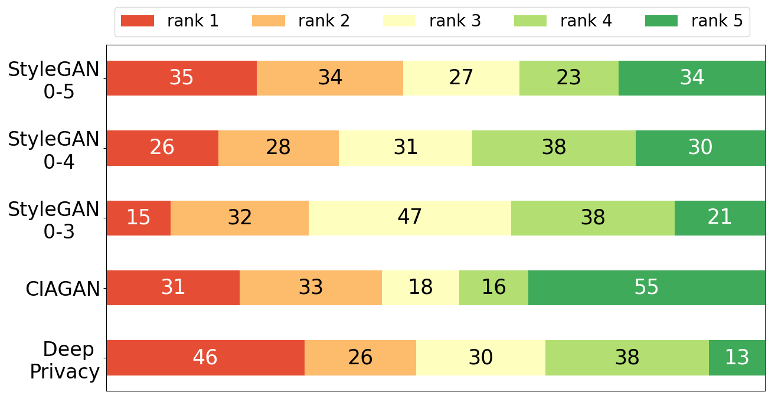}
  \caption{Naturalness of generated faces.  Participants found DeepPrivacy-generated faces to be more natural, followed by StyleGAN 0-5.}
  \label{fig:naturalness}
\end{figure}

Participants were asked to rank different faces generated by the 5 de-identification methods, especially if having a natural de-identified image is desired. Figure~\ref{fig:naturalness} shows the number of times each method was chosen to have a particular rank. Deep Privacy was chosen the most as the top choice (46 times) followed by StyleGAN 0-5 (35 times) while StyleGAN 0-3 was the least chosen (15 times). The means and standard error were 2.64 and 0.11 for Deep Privacy, 2.92 and 0.12 for StyleGAN 0-5, 3.12 and 0.10 for StyleGAN 0-3, 3.12 and 0.11 for StyleGAN 0-4, and 3.20 and 0.13 for CIAGAN. 
To examine H8, we ran five t-tests, comparing the mean of StyleGAN 0-5 with other methods, while the Bonferroni corrected significant level is 0.001. Our results showed that there is no statistically significant difference between the mean ranking of StyleGAN 0-5 compared with any other model, rejecting H8. 

\subsubsection{De-identification Method Preference}
The last survey question was about their preferred method of utility-preserving face obfuscation, where they could only choose one method. 
Overall, DeepPrivacy was the most liked method, with about one-third (50 participants) preferring this method. 
CIAGAN, StyleGAN 0-3, StyleGAN 0-4, and StyleGAN 0-5 had 40, 23, 22, and 18 votes, respectively. 
We used the open coding processing method to categorize the explanations. 
The explanations could be categorized into four major groups: \emph{utility preserving} (84 times), \emph{being natural or normal} (40 times), \emph{better in privacy} (35), \emph{quality and aesthetics} (10 times), \emph{not related or proper reasons} (24 times), and \emph{opposite reasons} (3 times). By opposite, we mean reasons that technically are negative but given to support their choices, like “looking unnatural,” “not realistic,” and “different races.” 
In 14 explanations, participants mentioned their choice maintained a similar facial expression. The number of mentions of the pose, gender, age, race or skin tone, and hair color were 7, 4, 2, 2, and 4 times, respectively. Compared to StyleGAN 0-5, for which the reasons were focused on “the same,” “the closest to the original,” and “the most similar,” the reasons for StyleGAN 0-4 were more about “close but not the same,” “almost the same but different race,” etc.  
Thirty-two (64\%) of those who selected DeepPrivacy indicated that they preferred it because it is the most similar to the original image maintaining image features and utility the most. This number was 15 (68\%),  11 (61\%), 18 (45\%), 8 (35\%) for StyleGAN 0-4, StyleGAN 0-5, CIAGAN, and StyleGAN 0-3, respectively. Nine (50\%) of the participants explained that they chose StyleGAN 0-5 because the de-identified faces are natural or normal. This number was 9 (39\%), 6 (27\%), 9 (22\%), and 5 (12.5\%) for StyleGAN 0-3, StyleGAN 0-4, Deep Privacy, and CIAGAN, respectively. 
Eight (35\%) of those who voted for StyleGAN 0-5 stated that because it was the best for privacy. This number was 7 (32\%), 5 (28\%), 9 (18\%), and 6 (15\%) for StyleGAN 0-4, StyleGAN 0-3, Deep Privacy, and CIAGAN.

\section{Discussion and Future work}
Our results show that in general, StyleGAN has better performance compared to DeepPrivacy and CIAGAN. 
The lower performance of DeepPrivacy and CIAGAN with respect to utility-preserving can be due to their emphasis on preserving specific attributes, such as pose and face shape, whereas in DeepPrivacy, only pose information is inserted into the generator and discriminator. 
In StyleGAN, however, there is no emphasis on a specific attribute, and depending on the mixing levels, it is possible to mix attributes of two faces.  
In terms of privacy, DeepPrivacy and CIAGAN both wipe out the face rectangle from the face and used the remaining as the input to the generator. This step excludes the identity information providing that the rectangle is detected properly. However, this might not be done properly, and the face recognition system takes advantage of it.  
Interestingly, even though StyleGAN does not remove the face entirely, our results showed that by inheriting fewer styles from the target, StyleGAN performs on par or better than DeepPrivacy and CIAGAN in protecting privacy. 

Employing face de-identification methods does not hide contextual features; therefore, using these methods alone might not protect users' privacy. Implementing attacks regarding context information is out of the scope of this paper. 
Using synthetic faces to replace all the faces in a dataset can positively and negatively affect downstream applications. The uncontrolled addition of synthetic data can cause unwanted behaviors of downstream models, e.g., adding to the bias. However, adding them in a controlled setting can help mitigate bias. For example, Kortylewski et al.~\cite{kortylewski2019analyzing} proposed training face recognition models using synthetic faces with different poses to reduce the damage of dataset bias in real data. 

In future work, we can examine the use of random vectors in StyelGAN to provide differential privacy and tune the privacy/ de-identification dial. We can also explore augmenting StyleGAN by imposing additional noise, to improve its privacy-preserving property. 
StyleGAN-generated faces can also be used by face-swapping methods. 
Classifiers can be developed to determine the target's face attributes, which helps to select a candidate to replace the target's face. One challenge of this approach is to generate faces for all combinations of features and poses. Moreover, it is not trivial to obtain a new face with the same quality as the original image that can be smoothly dissolved into the original background.
\section{Conclusion}
\label{sec:conclusion}

StyleGAN was initially proposed for generating imaginary faces. Our findings showed that it is also a great candidate for face de-identification. 
While most existing works focus on evaluating face de-identification methods against either machine learning attacks or human observers, this work evaluated StyleGAN against all, considering seven different attack models. Moreover, for the first time, we systematically evaluated and compared the utility-preserving properties of StyleGAN, DeepPrivacy, and CIAGAN. 
Our experiments show that StyleGAN with mixing levels 0-5, 0-4, and 0-3 protects privacy on par or even better than DeepPrivacy and CIAGAN. 
There is a trade-off between privacy and utility. When privacy against machine learning models has a higher priority, then StyleGAN with lower mixing levels, such as 0-2 or 0-3, can be employed and when utility is more important, then StyleGAN with higher mixing levels, e.g., 0-4 and 0-5 can be used.  
Interestingly, we found DeepPrivacy outperforming CIAGAN in almost all the experiments, while the model is older. 
Our user study results illustrated that people have different preferences when picking a face de-identification method. Therefore, providing a face de-identification kit that allows selecting from these methods might encourage and increase data privacy practices among people.

\begin{acks}
This work is partially supported by the National Science Foundation under Award III Medium 2107296 at the University of Texas at Arlington.  
\end{acks}
\bibliographystyle{ACM-Reference-Format}
\bibliography{refs}

\appendix

\begin{figure*}[t]
% \Centering
\begin{center}
\includegraphics[width=0.85\textwidth]{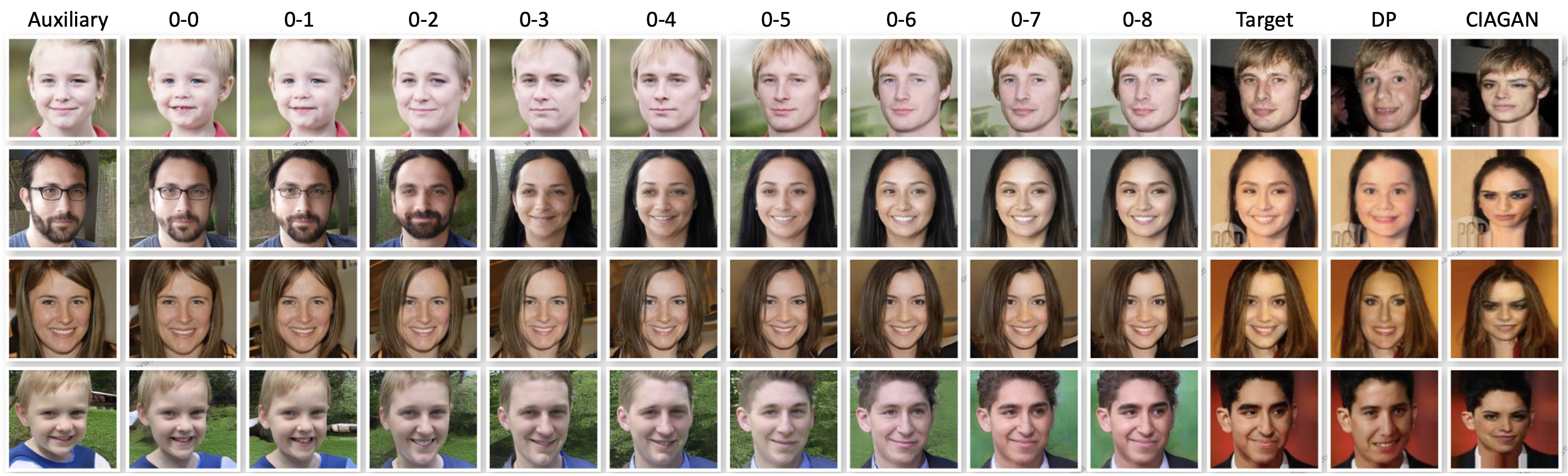}
\caption{De-identified images generated by StyleGAN 0-0 to 0-8, DeepPrivacy and CIAGAN. As more styles are inherited from the target, the generated face will be more like the target and less like the auxiliary face.}
\label{fig:cvpr}
\end{center}
\end{figure*}
% %%%%%%%%%%%%%%%%%%%%%%%%%%%%%%%%%%%%%%%%%%%%%%%%%%%%%%%%%%%%%%%%%%%%%%%%%%%%%%
\begin{figure}[t]
  \includegraphics[scale=.45]{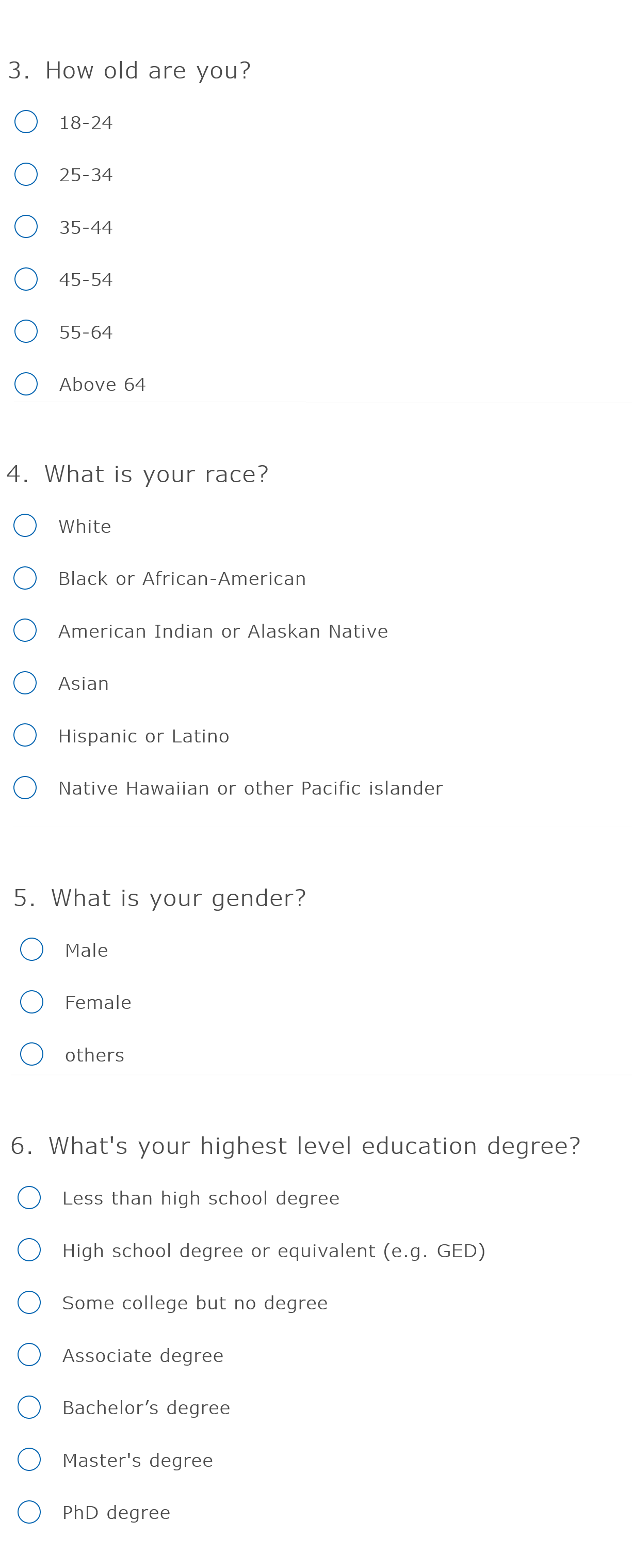}
  \caption{Demographics questions}
  \label{fig:demographicsq.png}
\end{figure}
%%%%%%%%%%%%%%%%%%%%%%%%%%%%%%%%%%%%%%%%%%%%%%%%%%%%%%%%%%%%%%%%%%%%%%%%%%%%
\section{Survey Questions}
\label{sec:set-diff-dodis}
\begin{figure}[t]
  \includegraphics[width=\columnwidth]{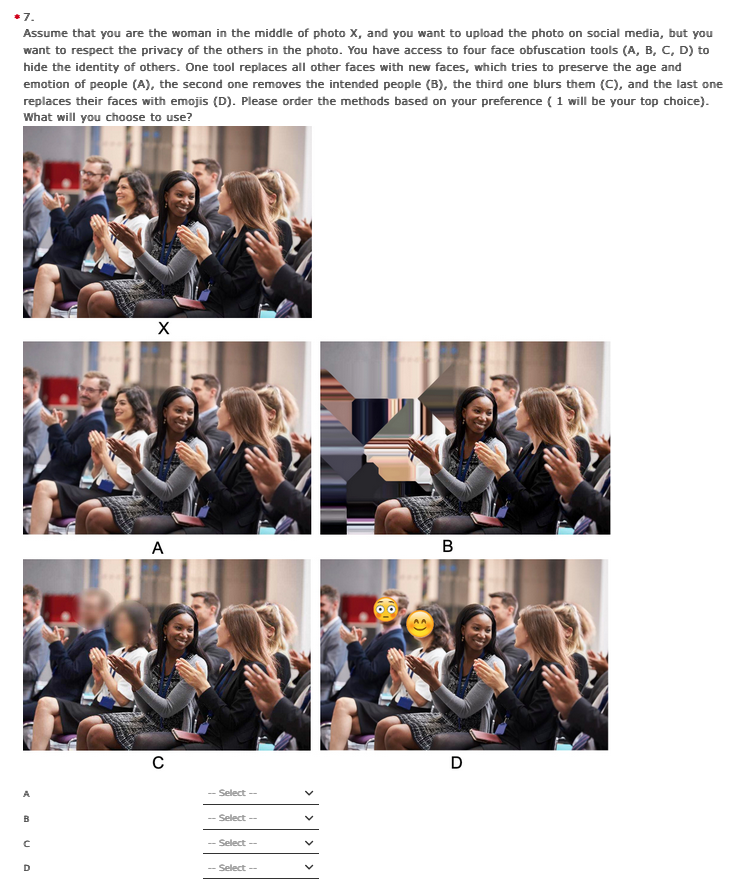}
  \caption{Obfuscation method preference}
  \label{fig:survey_tool_rank.png}
\end{figure}

\begin{figure}[t]
  \includegraphics[width=\columnwidth]{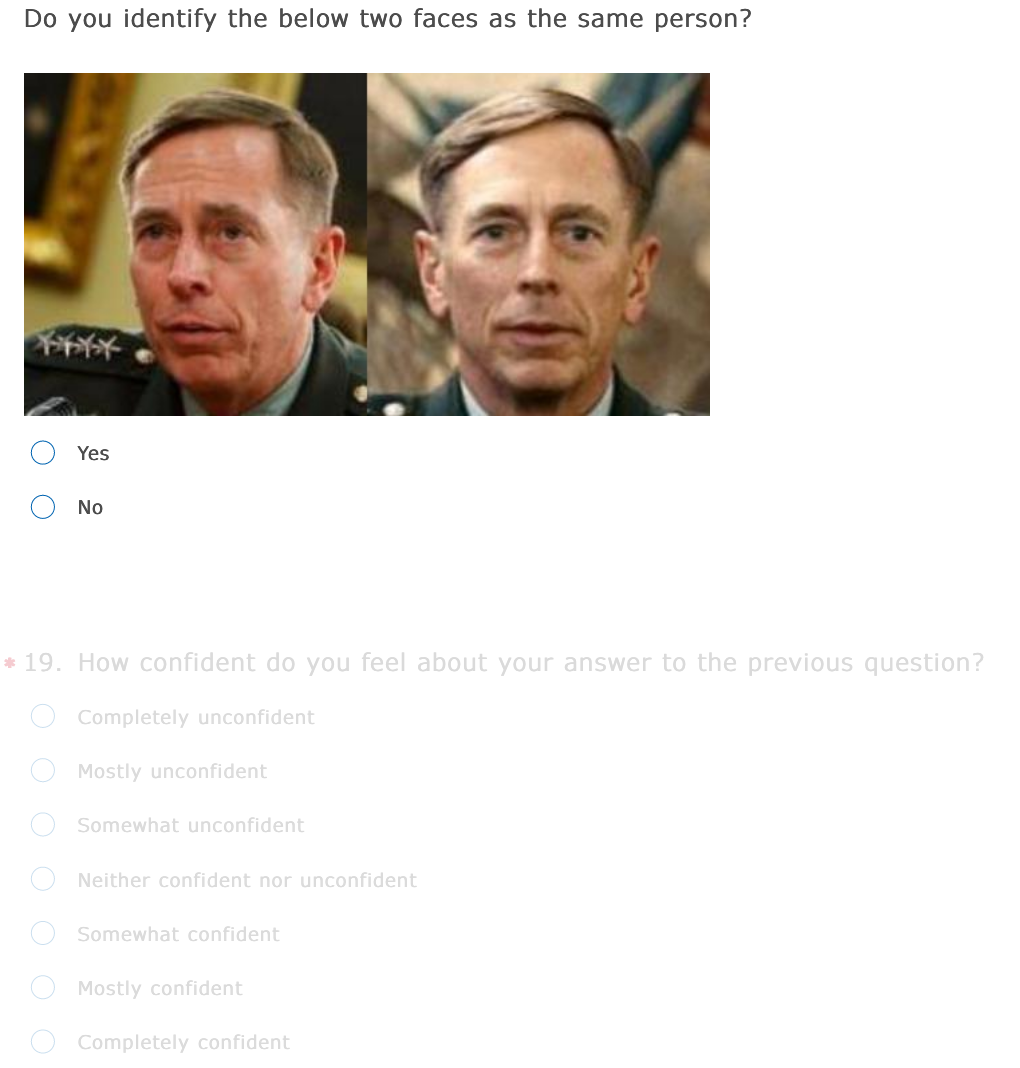}
    \caption{ A sample for the verification question for As Is condition when the target is present }
  \label{fig:survey_verification_asis.png}
\end{figure}

A screenshot of the demographic questions is shown in Figure~\ref{fig:demographicsq.png}. 
Figure~\ref{fig:survey_tool_rank.png} shows the question used to order different obfuscation methods. 
% %%%%%%%%%%%%%%%%%%%%%%%%%%%%%%%%%%%%%%%%%%%%%%%%%%%%%%%%%%%%%%%%%%%%%%%%%%%%%%
%%%%%%%%%%%%%%%%%%%%%%%%%%%%%%%%%%%%%%%%%%%%%%%%%%%%%%%%%%%%%%%%%%%%%%%%%%%%
Figure~\ref{fig:survey_verification_asis.png} shows a sample verification question for the As Is condition followed by a question about the confidence level when answering the verification question. 

\begin{figure}[t!]
\center
\includegraphics[width=0.7\columnwidth]{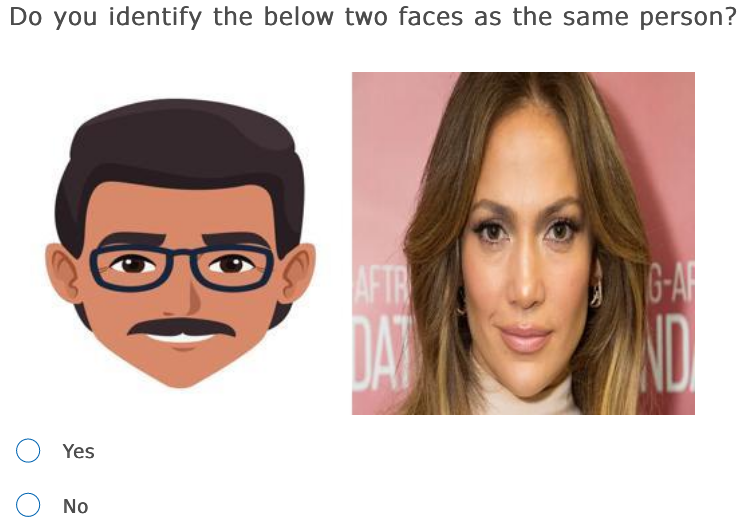}
\caption{Attention check question for the verification section when two faces are obviously different}
\label{fig:survey_attention_check1.png}
\end{figure}

\begin{figure}[t!]
\center
\includegraphics[width=0.7\columnwidth]{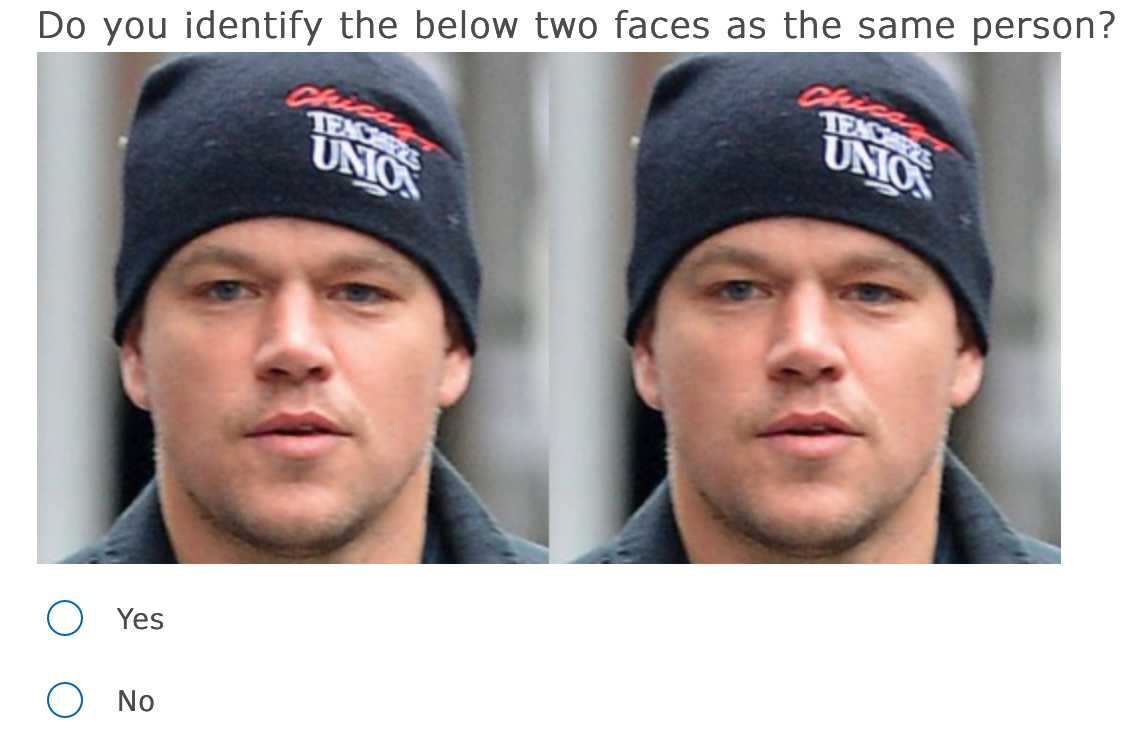}
\caption{Attention check question for the verification section when two faces are the same}
\label{fig:survey_attention_check2.png}
\end{figure}

Two attention-checking questions are shown in Figures~\ref{fig:survey_attention_check1.png} and \ref{fig:survey_attention_check2.png}, which expect obvious "no" and "yes" respectively. 
%%%%%%%%%%%%%%%%%%%%%%%%%%%%%%%%%%%%%%%%%%%%%%%%%%%%%%%%%%%%%%%%%%%%%%%%%%%%%%
%%%%%%%%%%%%%%%%%%%%%%%%%%%%%%%%%%%%%%%%%%%%%%%%%%%%%%%%%%%%%%%%%%%%%%%%%%%%
%%%%%%%%%%%%%%%%%%%%%%%%%%%%%%%%%%%%%%%%%%%%%%%%%%%%%%%%%%%%%%%%%%%%%%%%%%%%

\begin{figure}[t!]
\center
\includegraphics[width=\columnwidth]{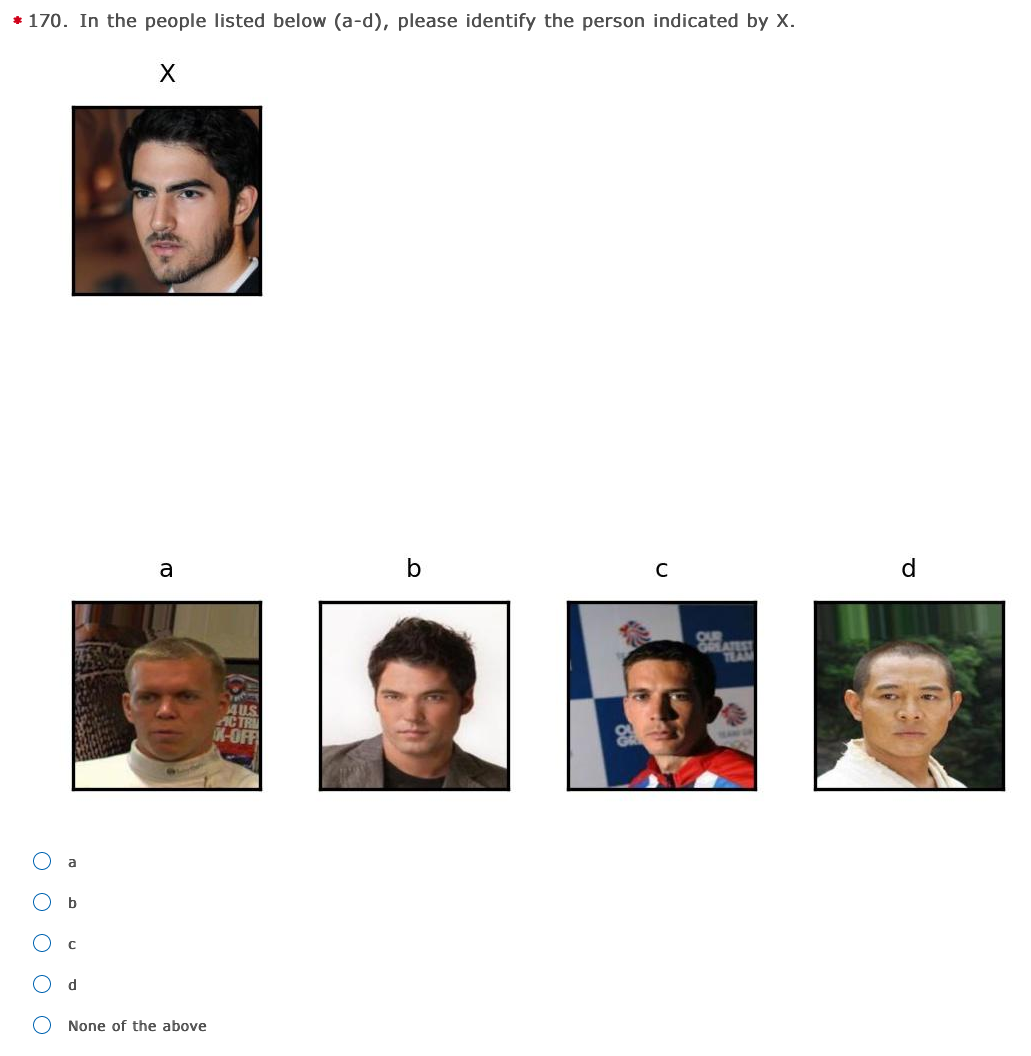}
\caption{Identification Question Sample}
\label{fig:survey_identification.png}
\end{figure}

A sample identification question using StyleGAN 0-4 as the identification method is given in Figure~\ref{fig:survey_identification.png}. 

%%%%%%%%%%%%%%%%%%%%%%%%%%%%%%%%%%%%%%%%%%%%%%%%%%%%%%%%%%%%%%%%%%%%%%%%%%%%

\begin{figure}[t!]
% \lefting
\includegraphics[width=\columnwidth]{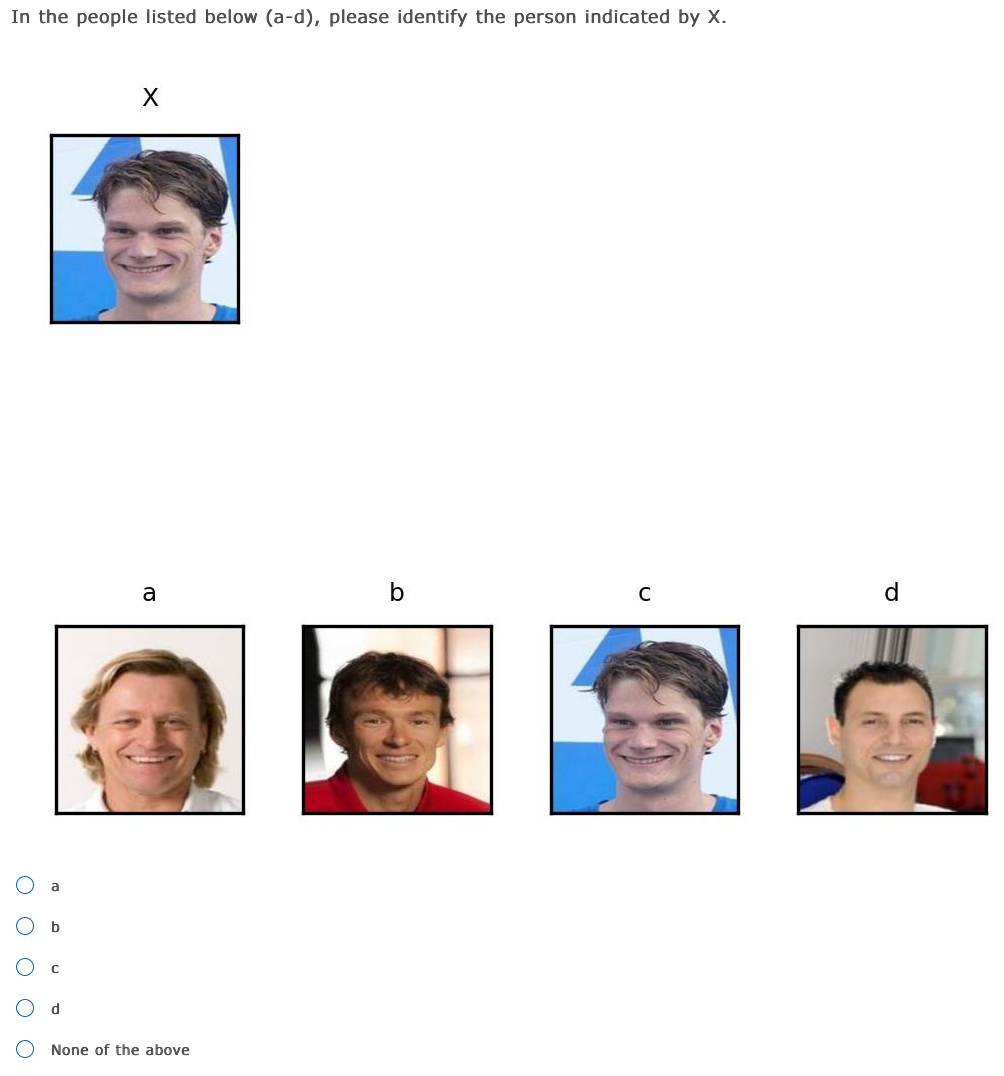}
\caption{The attention-checking question for the identification section.}
\label{fig:survey_identification_attention_check.png}
\end{figure}

The attention-check question of the identification section is shown at Figure~\ref{fig:survey_identification_attention_check.png} which requires choosing option "c". 

\begin{figure}[t!]
\center
\includegraphics[width=\columnwidth]{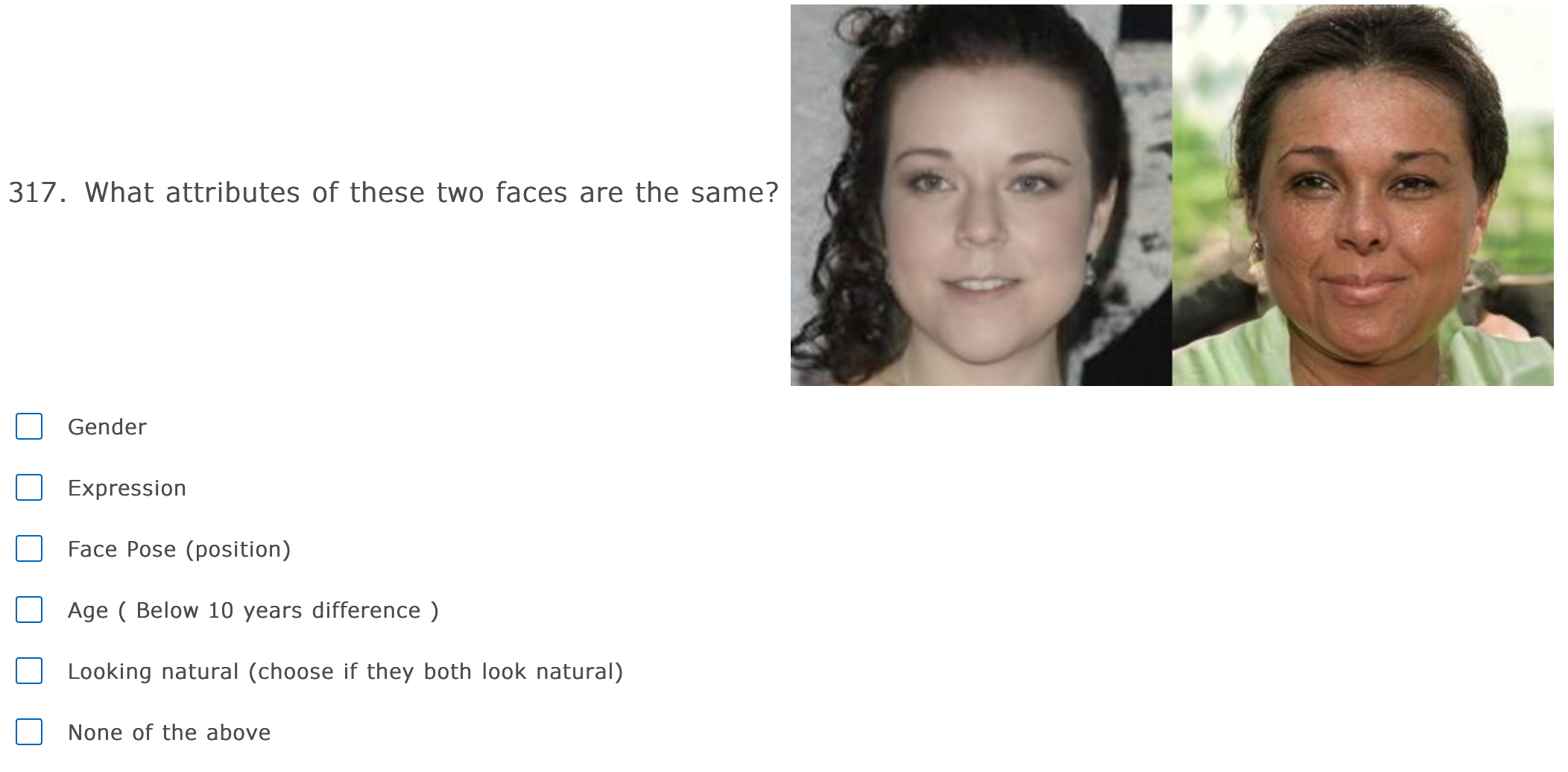}
\caption{Utility Preserving Question Sample}
\label{fig:survey_utility.png}
\end{figure}

%%%%%%%%%%%%%%%%%%%%%%%%%%%%%%%%%%%%%%%%%%%%%%%%%%%%%%%%%%%%%%%%%%%%%%%%%%%%
A sample where two photos are shown and asked if they share the same facial attributes or not is shown in Figure~\ref{fig:survey_utility.png}. 

\begin{figure}[t!]
\center
\includegraphics[width=\columnwidth]{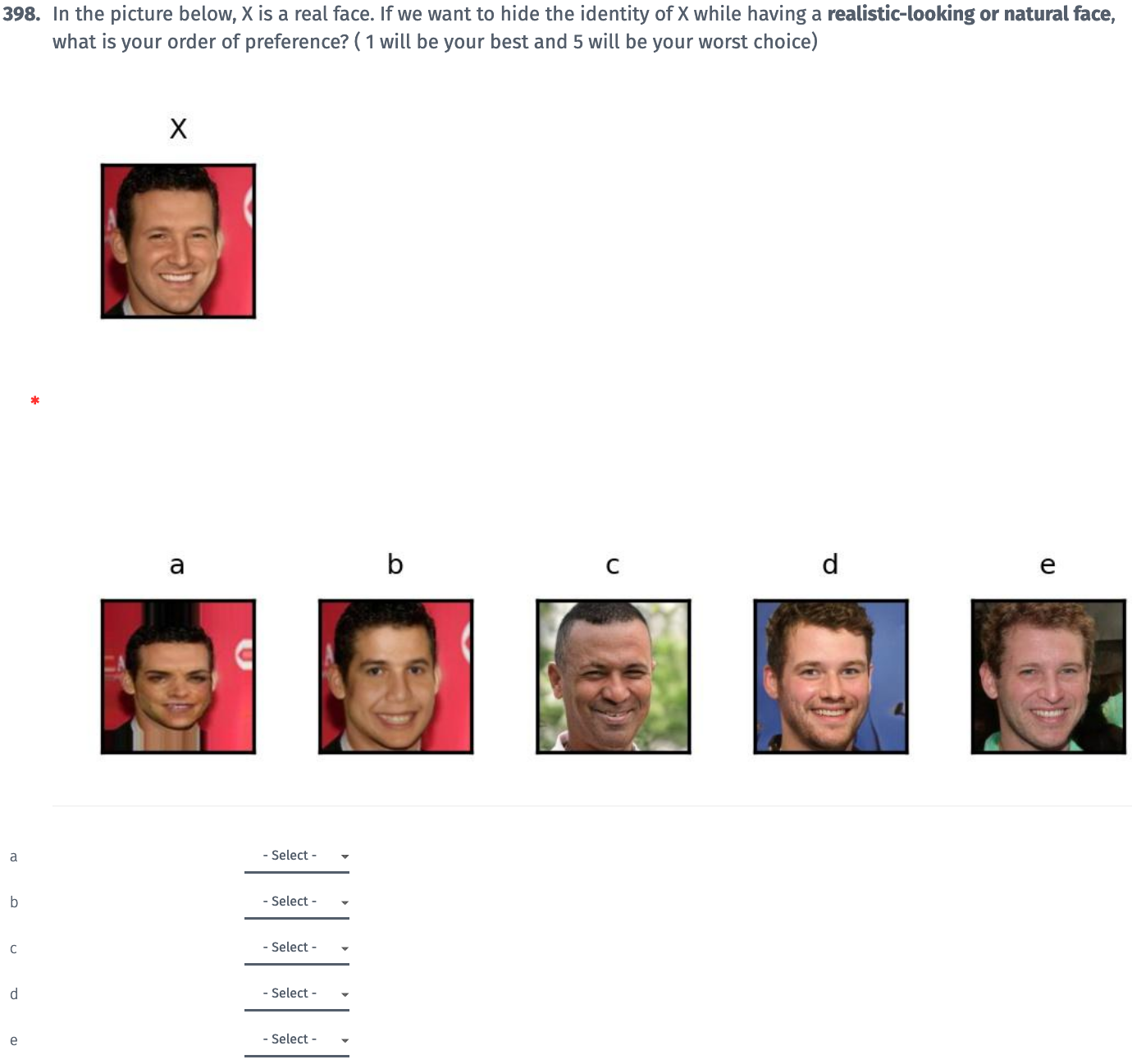}
\caption{A question sample for natural obfuscation method ranking}
\label{fig:survey_preference_rank.png}
\end{figure}

\begin{figure}[t!]
\center
\includegraphics[width=\columnwidth]{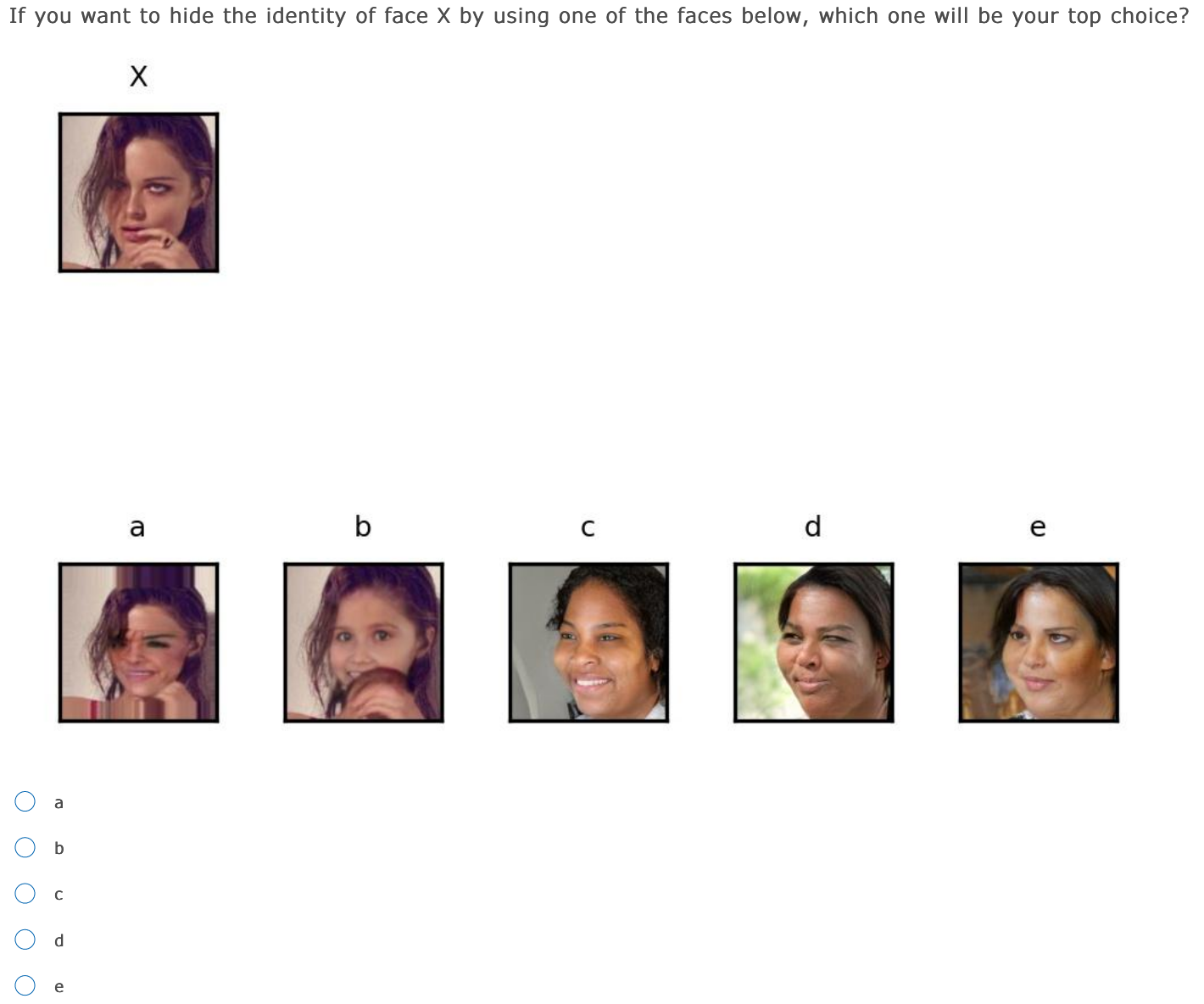}
\caption{A sample for utility-preserving obfuscation method preference question}
\label{fig:survey_preference.png}
\end{figure}

A question sample for natural obfuscation method ranking is shown in Figure~\ref{fig:survey_preference_rank.png}. Finally, the last question which shows a target face and asks to choose the most favorite utility-preserving de-identified version using CIAGAN, Deep Privacy, StyleGAN 0-3, StyleGAN 0-4, and StyleGAN 0-5 respectively is shown in Figure~\ref{fig:survey_preference.png}.

\end{document}